\documentclass[11pt]{article}
\PassOptionsToPackage{table}{xcolor}
\usepackage[preprint]{acl}

\usepackage{times}
\usepackage{latexsym}
\usepackage[T1]{fontenc}
\usepackage[utf8]{inputenc}
\usepackage{microtype}
\usepackage{inconsolata}

\usepackage{amsmath,amsfonts,bm}

\def\eqref#1{equation~\ref{#1}}

\def\1{\bm{1}}

\DeclareMathAlphabet{\mathsfit}{\encodingdefault}{\sfdefault}{m}{sl}
\SetMathAlphabet{\mathsfit}{bold}{\encodingdefault}{\sfdefault}{bx}{n}

\usepackage{url}
\usepackage{booktabs}
\usepackage{multirow}
\usepackage{graphicx}
\usepackage{amssymb}
\usepackage{makecell}
\usepackage{algorithm}
\usepackage{wrapfig}
\usepackage{enumitem}
\usepackage{subcaption}
\usepackage{algpseudocode}
\usepackage[skins]{tcolorbox}

\newcommand{\deltattg}{{\itshape DeltaTTG}}
\newcommand{\deltattgvar}[1]{\deltattg{}\textsubscript{{\bfseries\itshape #1}}}

\title{\deltattg: Training Multi-Turn Agents with a Token-to-Go Progress Meter}

\author{
  Xuyan Ye\textsuperscript{12}\thanks{Equal contribution.\quad\textsuperscript{\textdagger}Corresponding author.\protect\\\textsuperscript{\textdaggerdbl}Work done during an internship at NTU.}\textsuperscript{\textdaggerdbl},
  Xiyan Fu\textsuperscript{1*},
  Shengda Fan\textsuperscript{2*},
  Jiawei Cui\textsuperscript{1},\\
  \textbf{Yankai Lin\textsuperscript{2},
  Wei Lu\textsuperscript{1\textdagger}}\\[0.5ex]
  \textsuperscript{1}College of Computing and Data Science, Nanyang Technological University\\
  \textsuperscript{2}Gaoling School of Artificial Intelligence, Renmin University of China\\[0.5ex]
  \texttt{yexvyan0923@ruc.edu.cn},\quad\texttt{wei.lu@ntu.edu.sg}
}

\begin{document}

\begingroup
\makeatletter
\long\def\@makefntext#1{\noindent\@makefnmark#1}
\maketitle
\makeatother
\endgroup

\begin{abstract}
Reinforcement learning with verifiable rewards (RLVR) for multi-turn agents relies on sparse terminal rewards that say little about individual steps. Within-group step comparisons depend on reliable state matching and sufficiently diverse sampled outcomes, while direct step scoring may require costly external evaluators, human annotations, or task-specific heuristics. We introduce \textbf{\deltattg{}} to learn a reusable progress signal through a shared progress meter, using low-cost supervision from diverse policies' recorded trajectories without task-specific progress rules. Its supervision target, \textbf{Token-to-Go (TTG)}---the number of tokens remaining to be generated until task completion---serves as an observable proxy for remaining completion effort. During actor training, changes in the frozen meter's estimates redistribute outcome-derived advantages across steps without altering the reward, preserving each trajectory's advantage sign and token-average scale. On ALFWorld, PaperSearchQA, and BFCLv3, \deltattg{} improves the average score over GRPO by 5.7 and 7.2 points with Qwen3-4B and Llama-3.1-8B-Instruct, respectively, outperforms all compared RL baselines on average, and keeps actor-training GPU-hours close to GRPO. Further experiments show improved sample efficiency and out-of-domain performance, and support the meter's ability to distinguish step quality and guide test-time action selection across actor sizes. The code is available at \url{https://github.com/Xuyan923r/DeltaTTG}.
\end{abstract}

\begin{figure}[t]
    \centering
    \includegraphics[width=\columnwidth]{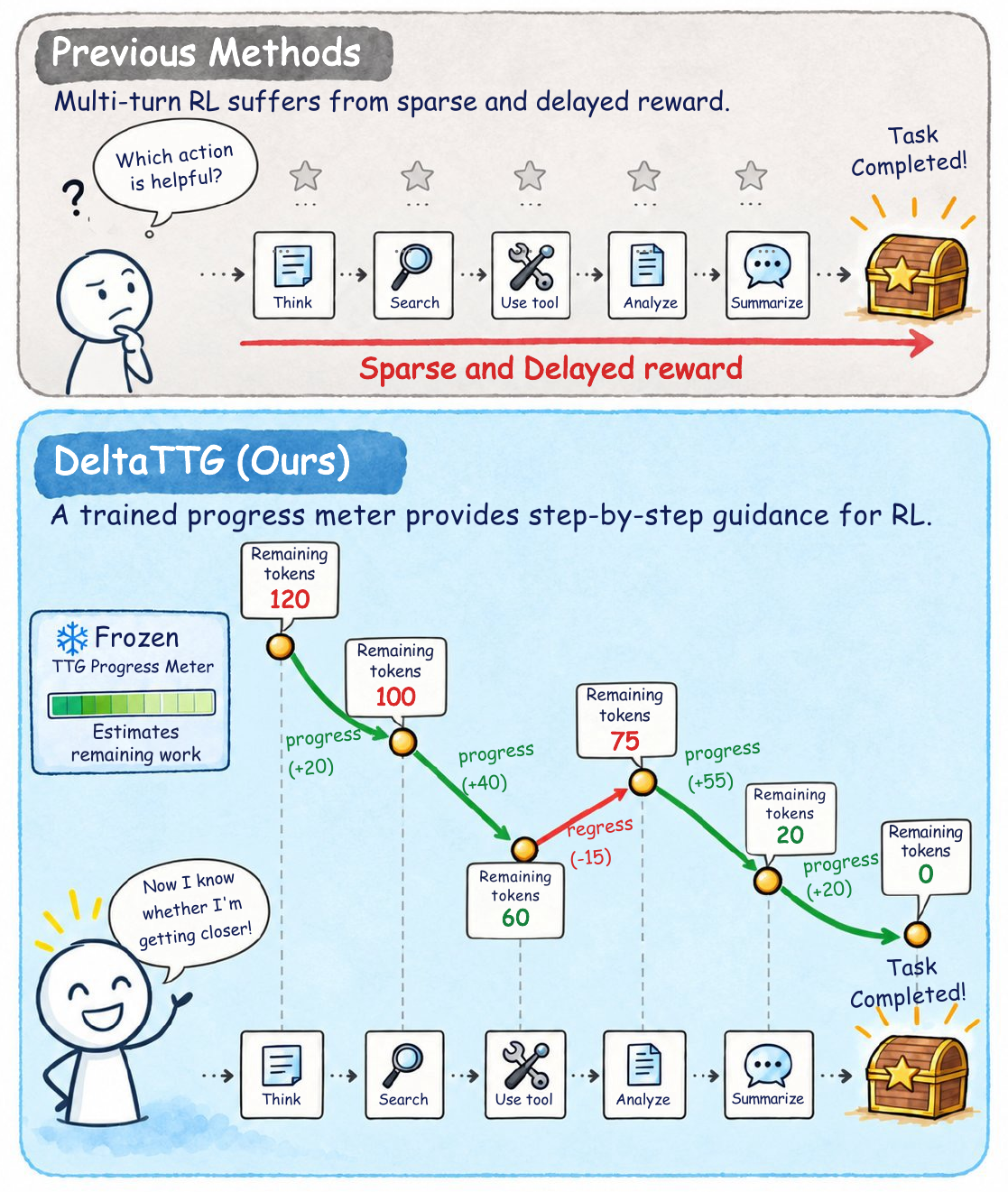}
    \caption{RLVR for agents faces sparse and delayed rewards. \deltattg{} learns a progress meter from diverse trajectories, measures each transition's change in estimated remaining work, and uses it to redistribute the trajectory's outcome credit.}
    \label{fig:teaser}
\end{figure}

\section{Introduction}
\label{sec:intro}

Large language models are increasingly trained as agents that solve tasks through multi-turn interaction with tools and environments~\citep{liu2024agentbench,zhou2024webarena,jin2025search,da2025agent,wang2023voyager}, often with reinforcement learning from verifiable outcomes~\citep{shao2024deepseekmath}. Such rewards are \textbf{sparse and delayed}: a single terminal signal arrives after a long sequence of decisions~\citep{wang2025ragen,luo2025agent} and says little about which steps helped or hurt, making step-level credit assignment difficult.

Recent methods derive step-level credit by comparing actions within the current rollout group~\citep{feng2026group,he2026hierarchy}. However, the quality of these comparisons depends on both the comparability of grouped states and the evidence available in the sampled rollouts. In open-ended interactions, exact state matches can be limited, and similarity-based grouping can introduce noise by pooling states with different decision-relevant conditions; limited outcome diversity and confounding later decisions further weaken local comparisons. Methods that score steps directly may instead require costly judges, human step annotations, additional rollouts~\citep{zheng2023judging,lightman2024let,choudhury2025process,zhao2026genprm,kazemnejad2024vineppo}, or task-specific heuristics~\citep{wang2026information,wang2026milestone}. These challenges raise a question: \textit{Can we learn a reusable progress signal from ordinary agent trajectories that evaluates each action beyond the evidence in the current rollout group?}

A learned value model is a natural candidate, but it is tied to the actor whose rollouts it was trained on, and under sparse rewards many states share the same return target, flattening its estimates~\citep{li2026rethinking}. We instead measure progress as \textbf{the work remaining to complete the task}, using \textbf{Token-to-Go (TTG)}---the number of tokens still to be generated before success---as an observable proxy. Our core hypothesis is that, although remaining token usage varies with model style and interaction harness, beneficial actions from the same state leave fewer tokens to completion on average across many trajectories than harmful ones (e.g., each ticket tier an agent creates removes part of the remaining work; Appendix~\ref{app:case_study}), with failed continuations assigned a large remaining-work cost. TTG labels come at no extra cost from recorded trajectories, and pooling trajectories from models of different families, sizes, and capability levels yields a shared estimate rather than one actor's behavior.

Building on this idea, we introduce \textbf{\deltattg{}}, which trains \textbf{a reusable progress meter} on such trajectories; one meter per domain serves actors of different families and sizes without retraining. During actor training, the frozen meter scores each state, and the change in its estimate between consecutive states determines how the trajectory's outcome advantage is distributed across steps (Figure~\ref{fig:teaser}, bottom).

We evaluate \deltattg{} across two model families and parameter scales on ALFWorld, PaperSearchQA, and BFCLv3, covering interactive decision making, information seeking, and tool use. Our contributions are threefold: (1) we propose Token-to-Go as a low-cost, observable supervision target for task progress, extracted from recorded trajectories; (2) we introduce \deltattg{}, which uses a frozen progress meter to redistribute outcome-derived advantages across steps without adding rewards; and (3) we show gains over strong RL baselines with improved sample efficiency and out-of-domain performance, at actor-training cost close to GRPO, and validate the meter itself through step-quality assessment, test-time action guidance across actor sizes, and cross-domain transfer.

\begin{figure*}[t]
    \centering
    \includegraphics[width=\textwidth]{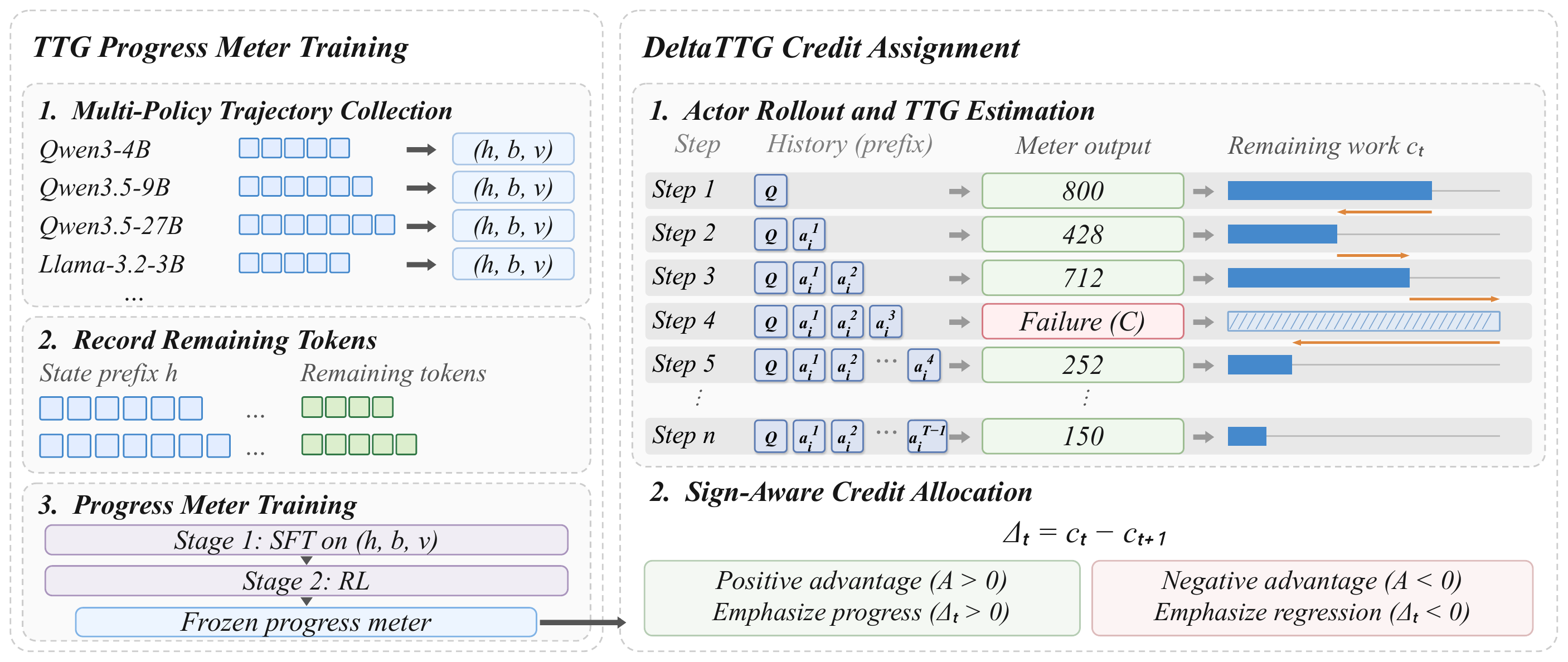}
    \caption{
    Overview of the \deltattg{} framework. A TTG progress meter is trained on recorded agent trajectories through SFT and RL, using observed continuation outcomes and remaining-token labels. During actor training, the frozen meter estimates remaining completion effort; changes between successive states guide step-level redistribution of outcome-derived advantages, providing fine-grained process supervision.
    }
    \label{fig:deltattg_overview}
\end{figure*}

\section{Related Work}

\paragraph{Reinforcement Learning for Language Agents.}
Reinforcement learning strengthens several capabilities of LLM-based agents, including multi-turn decision making through environment interaction~\citep{zhou2024archer,wang2025ragen,luo2025agent}. It improves browser navigation~\citep{qi2025webrl}, search-query formulation and evidence integration~\citep{jin2025search}, and repository-level software problem solving~\citep{da2025agent,fan2026darc}. Beyond task execution, jointly optimizing reasoning and memory consolidation helps agents sustain long interactions with compact context~\citep{zhou2026mem1}, while failure diagnosis, corrective feedback, and learning from repaired actions support agent self-correction~\citep{zhu2025llm,zhu2026aed}.

\paragraph{Process Supervision and Credit Assignment.}
Existing approaches to process supervision can be broadly grouped into two categories. The first provides direct supervision for process units through external LLM evaluators, human step annotations, or verifier feedback~\citep{lightman2024let,zhao2026genprm,liu2026agentic,kim2026verifiableprocess,li2026vict,li2026verifiedcritical}. It also draws on search or reference reasoning~\citep{zhang2024rest,zhong2026p2s}, task-specific heuristics such as information gain and milestones~\citep{wang2026information,lu2026abseeker,wang2026milestone,zhang2026trca}, and on-policy distillation, which obtains dense token-level supervision from teachers on student-generated rollouts~\citep{zhu2026many}. The second redistributes outcome-based credit across process units~\citep{fan2026tips,wang2025sparl,zhang2026vzero}. Examples include constructing relative step advantages by comparing sampled returns under shared states, histories, or transition dynamics~\citep{feng2026group,he2026hierarchy,liu2026shadow}, and learning step contributions that reconstruct the final outcome~\citep{wang2025sparl}. \deltattg{} learns remaining-work estimates from cross-policy Token-to-Go labels and uses changes in these estimates to redistribute outcome-derived advantages, without matched peer states or additional process rewards.

\section{Methodology}
\label{sec:method}

As illustrated in Figure~\ref{fig:deltattg_overview}, \deltattg{} first learns a shared TTG progress meter from recorded trajectories (Section~\ref{sec:predictor_training}) and then uses its frozen estimates for actor credit assignment (Section~\ref{sec:deltattg}).

\subsection{Training a Token-to-Go Progress Meter}
\label{sec:predictor_training}
\label{sec:ttg_construction}

\paragraph{Token-to-Go as supervision for task progress.}
For a recorded trajectory $\tau_i^{\mathrm{ref}}$ with $T_i^{\mathrm{ref}}$ actions, let $h_{i,t}^{\mathrm{ref}}$ denote the history before action $t$ and $n_{i,u}^{\mathrm{ref}}$ the number of assistant-generated tokens in action $u$. We define its recorded Token-to-Go as
\begin{equation}
\mathrm{TTG}_{i,t}=\sum_{u=t}^{T_i^{\mathrm{ref}}-1}n_{i,u}^{\mathrm{ref}}.
\label{eq:ttg_definition}
\end{equation}
Building on BAGEN's prefix-based remaining-budget formulation~\citep{lin2026bagen}, we measure completion effort in assistant-generated tokens, including reasoning, tool calls, and final answers. Environment observations remain part of the input context but are excluded from this count, as are tokens already in the prefix. For a successful trajectory, the tokens needed to complete the task from a given state depend on the policy, interaction harness, and subsequent decisions. Each recorded TTG therefore provides one observation of remaining completion effort.

\paragraph{Constructing supervision from recorded trajectories.}
We split trajectories from diverse policies at assistant-action boundaries into observable prefixes and recorded suffixes. Each prefix receives an observed completion label and, for successful trajectories, a TTG target:
\begin{equation}
\begin{aligned}
b_{i,t}&=\mathbf{1}[\tau_i^{\mathrm{ref}}\text{ succeeds}],\\
v_{i,t}&=
\begin{cases}
\mathrm{TTG}_{i,t},&b_{i,t}=1,\\
\mathrm{null},&b_{i,t}=0.
\end{cases}
\end{aligned}
\end{equation}
Failure labels describe recorded outcomes, not intrinsic unsolvability. Pooling these observations across policies trains a shared task progress meter to estimate remaining completion effort, rather than any particular actor's exact future token consumption.

\paragraph{SFT for structured progress outputs.}
The meter returns $f_\phi(h)=(\hat b,\hat v)$: a binary completion estimate and a nonnegative remaining-work estimate learned from recorded continuations across the training policies. These shared estimates are then applied to the current actor's histories, with $\hat v=\mathrm{null}$ when $\hat b=0$. Given a prefix $h$ and its structured target answer $y$, SFT minimizes
\begin{equation}
\mathcal L_{\mathrm{SFT}}(\phi)=-\mathbb E_{(h,y),j}\!\left[\log p_\phi(y_j\mid h,y_{<j})\right].
\label{eq:meter_sft}
\end{equation}
The expectation is over answer tokens only. This stage teaches the output format and an initial mapping from prefixes to the recorded labels.

\paragraph{RL for refining progress estimates.}
We refine the meter with GRPO~\citep{shao2024deepseekmath}. For a labeled prefix $(h,b,v)$, a sampled answer $y$ receives
\begin{equation}
r_{\mathrm{pred}}(y;b,v)=
\begin{cases}
0.2+0.8u(\hat v,v),&\hat b=b=1,\\
1,&\hat b=b=0,\\
-1,&\text{otherwise},
\end{cases}
\label{eq:meter_reward}
\end{equation}
where invalid outputs fall under the last case. For correctly predicted successes, token accuracy is
\begin{equation}
u(\hat v,v)=\max\!\left(0,1-\frac{|\hat v-v|}{\max(|\hat v|,|v|,1)}\right).
\label{eq:meter_token_accuracy}
\end{equation}
This reward encourages correct completion forecasts and accurate remaining-work estimates. We normalize sampled rewards by their within-group mean and standard deviation for GRPO updates, then freeze the trained meter during actor optimization (details in Appendix~\ref{app:progress_meter_training}; evaluation in Section~\ref{sec:analysis}).

\subsection{\deltattg{} Credit Assignment}
\label{sec:deltattg}

\paragraph{Rollout and outcome advantages.}
For each task, the rollout policy samples $G$ trajectories. Trajectory $\tau_i$ contains $T_i$ actions; $h_{i,t}$ denotes the history before action $a_{i,t}$. The next history $h_{i,t+1}$ includes $a_{i,t}$ and the resulting environment observation. Let $n_{i,t}$ be the number of trainable tokens in that action and $N_i$ their total across the trajectory. Given terminal reward $R_i$, the GRPO~\citep{shao2024deepseekmath} advantage is
\begin{equation}
A_i=\frac{R_i-\mu_x}{\sigma_x+\epsilon}.
\label{eq:trajectory_advantage}
\end{equation}
Here, $\mu_x$ and $\sigma_x$ are computed over the task group after repeating each trajectory reward once per action (Appendix~\ref{app:group_advantages}). After collecting the full rollout batch, we score its decision histories with the frozen progress meter to obtain nonnegative credit weights with unit token mean within each trajectory.

\paragraph{Estimating transition progress.}
We use the frozen meter's remaining-work estimate as the state cost and assign a fixed failure-state cost $C>0$ to unsuccessful-completion estimates, distinguishing failure from low-effort completion. 

Terminal costs follow the observed outcomes:
\begin{equation}
\begin{aligned}
c_{i,t}
&=
\begin{cases}
\hat v_{i,t},
& \hat b_{i,t}=1,\\[4pt]
C,
& \hat b_{i,t}=0,
\end{cases}
\\
c_{i,T_i}
&=
\begin{cases}
0, & \text{success},\\
C, & \text{failure or truncation}.
\end{cases}
\end{aligned}
\end{equation}

The credit signal of action $a_{i,t}$ is its reduction in predicted remaining work:
\begin{equation}
\Delta_{i,t}=c_{i,t}-c_{i,t+1}.
\end{equation}
This difference of the progress potential $\Phi(h_{i,t})=-c_{i,t}$ measures estimated progress ($\Delta_{i,t}>0$) or regression ($\Delta_{i,t}<0$). We set $C$ to a large, dataset-dependent constant. Between nonterminal states, its value affects the estimated change in remaining work only when the completion estimate switches between success and failure. Such predictions remain rare on successful trajectories,\footnote{\label{fn:failure_cost_audit}In our AgentWorldModel training audit, only 6.3\%--10.5\% of successful trajectories contain any state that the meter judges unsuccessful.} with no clear increasing or decreasing trend over training updates. Thus, most successful trajectories contain no nonterminal transitions whose progress estimates depend on $C$.

\paragraph{Sign-aware credit allocation.}
We give estimated progress more positive credit when $A_i>0$ and assign estimated regression negative credit of greater magnitude when $A_i<0$. The nonnegative allocation mass is
\begin{equation}
m_{i,t}=
\begin{cases}
\max(\Delta_{i,t},0), & A_i>0,\\
\max(-\Delta_{i,t},0), & A_i<0,\\
1, & A_i=0.
\end{cases}
\end{equation}

\paragraph{Combining trajectory advantages with step-level progress.}
Mixing progress-based shares with a uniform share over steps gives
\begin{equation}
q_{i,t}=\frac{\rho}{T_i}+(1-\rho)\frac{m_{i,t}}{\sum_s m_{i,s}},
\end{equation}
where $\rho\in[0,1]$ retains credit for other steps and reduces sensitivity to progress-estimation errors. If $\sum_s m_{i,s}=0$, we use $q_{i,t}=1/T_i$.\footnote{Trajectories without any state estimated as completable, and groups with an invalid meter output, keep their original advantages (Algorithm~\ref{alg:deltattg}).}

We broadcast each step's weight to its trainable tokens and normalize the multipliers within each trajectory:
\begin{equation}
\widetilde A_{i,t,k}=A_iw_{i,t}=A_i\!\left[\mathcal B_M\!\left(\{q_{i,s}\}_s;\{n_{i,s}\}_s\right)\right]_t.
\label{eq:token_credit_weights}
\end{equation}
The operator $\mathcal B_M$ normalizes weights to unit token mean, caps multipliers at $M\geq1$, and redistributes excess weight among uncapped tokens in proportion to their remaining capacity (Appendix~\ref{app:bounded_normalization}). Trajectory advantages determine the update direction, while step weights control its strength. We use a uniform-mixing coefficient of $\rho=0.2$ and a token-multiplier cap of $M=5$. The cap limits advantage amplification from large $C$-induced progress changes, helping mitigate sensitivity to $C$ and stabilize actor updates.

\paragraph{Actor optimization.}
The actor uses the vanilla clipped loss with token-mean reduction. Let $r_{i,t,k}(\theta)$ be the current-to-rollout policy ratio for token $k$ of action $a_{i,t}$ at its full prefix, and $\bar r_{i,t,k}=\operatorname{clip}(r_{i,t,k},1-\eta_{\mathrm{low}},1+\eta_{\mathrm{high}})$, with lower and upper clipping widths $\eta_{\mathrm{low}}=0.2$ and $\eta_{\mathrm{high}}=0.28$. The clipped surrogate is
\begin{equation}
\begin{aligned}
\mathcal L_{\Delta}(\theta)
&=-\frac{1}{\sum_i N_i}\sum_{i,t,k}\ell_{i,t,k},\\
\ell_{i,t,k}
&=\min\!\left(r_{i,t,k}\widetilde A_{i,t,k},
\bar r_{i,t,k}\widetilde A_{i,t,k}\right).
\end{aligned}
\label{eq:actor_clipped_loss}
\end{equation}
GRPO assigns $A_i$ to every token; \deltattg{} changes only the advantage allocation in this loss. We also apply dual clipping ($d=3$) to negative advantages and include the configured KL and entropy terms (Appendix~\ref{app:actor_objective}).

\section{Experiments}

\subsection{Experimental Setup}
\label{sec:experimental_setup}

\paragraph{Models.}
We use Qwen3-4B and Llama-3.1-8B-Instruct as actors. We separately train three progress meters for text-based interaction, information seeking, and complex engineering tasks, using trajectories from ALFWorld, PaperSearchQA, and AgentWorldModel, respectively. Each meter is initialized from Qwen3-4B-Instruct-2507, frozen during actor optimization, and reused across both actor backbones in its corresponding task setting. The models used to generate trajectories for progress-meter training are described in Appendix~\ref{app:meter_data_reuse}.

\paragraph{Baselines.}
Our main comparison includes five baselines: (1) \textbf{ReAct}~\citep{yao2022react}, which interleaves reasoning and actions through prompting; (2) \textbf{GRPO}~\citep{shao2024deepseekmath}, which computes trajectory-level group-relative advantages without a critic; (3) \textbf{DAPO}~\citep{yu2026dapo}, which combines asymmetric clipping, dynamic sampling, and token-level policy optimization; (4) \textbf{GiGPO}~\citep{feng2026group}, which derives step-level advantages by grouping actions at shared states; and (5) \textbf{HGPO}~\citep{he2026hierarchy}, which refines step-level grouping using historical context (settings in Appendix~\ref{app:baselines}). 

\paragraph{Benchmarks.}
Our main evaluation covers (1) \textbf{ALFWorld}~\citep{shridhar2020alfworld}, for interactive decision making; (2) \textbf{PaperSearchQA}~\citep{burgess2026papersearchqa}, for search and reasoning over scientific papers; and (3) \textbf{BFCLv3}~\citep{patil2025bfcl}, for function calling and multi-turn tool use. For the tool-use setting, we train on \textbf{AgentWorldModel (AWM)}~\citep{wang2026agentworldmodel} and evaluate on BFCLv3. We also assess cross-domain generalization on \textbf{BrowseComp-Plus}~\citep{chen2025browsecompplus}, for web information seeking; \textbf{Terminal-Bench 2.0}~\citep{merrill2026terminalbench}, for complex terminal tasks; and \textbf{$\boldsymbol{\tau^2}$-bench}~\citep{barres2025tau2bench}, for conversational tool use and user coordination (splits and protocols in Appendix~\ref{app:benchmarks}).

\begin{table*}[!t]

\centering

\caption{
Performance comparison on ALFWorld, PaperSearchQA, and BFCLv3 (\%, $\uparrow$). We report success rates on ALFWorld Seen/Unseen and PaperSearchQA Original/Paraphrased, and scores on the four BFCLv3 categories; Hall. denotes Hallucination/Irrelevance. For repeated evaluations, entries show the mean and population standard deviation over three test runs. Closed models provide reference results; within each open-backbone block, the highest and second-highest mean scores are \textbf{bold} and \underline{underlined}, respectively. Avg. is the mean of the three benchmark averages (ALFWorld, PaperSearchQA, and BFCLv3), reported without a standard deviation.
}
\label{tab:main_results}

\setlength{\tabcolsep}{5pt}
\renewcommand{\arraystretch}{1.08}

\resizebox{\textwidth}{!}{
\begin{tabular}{l|cc|cc|cccc|c}
\toprule

\multirow{2}{*}{\textbf{Method}}
&
\multicolumn{2}{c|}{\textbf{ALFWorld}}
&
\multicolumn{2}{c|}{\textbf{PaperSearchQA}}
&
\multicolumn{4}{c|}{\textbf{BFCLv3}}
&
\multirow{2}{*}{\textbf{Avg.}}
\\

\cmidrule(lr){2-3}
\cmidrule(lr){4-5}
\cmidrule(lr){6-9}

&
\textbf{Seen}
&
\textbf{Unseen}
&
\textbf{Original}
&
\textbf{Paraphrased}
&
\textbf{Non-Live}
&
\textbf{Live}
&
\textbf{Multi-Turn}
&
\textbf{Hall.}
&
\\

\midrule

\multicolumn{10}{l}{\textbf{\textit{Closed Models}}}
\\[-1pt]

GPT-5.6-Sol-low
& $81.43_{\pm 1.75}$ & $84.08_{\pm 0.93}$
& $62.56_{\pm 0.17}$ & $59.57_{\pm 0.38}$
& $76.60_{\pm 0.32}$ & $71.01_{\pm 0.24}$
& $49.71_{\pm 2.72}$ & $68.71_{\pm 0.59}$
& 70.11
\\

Claude-Opus-5-low
& $97.38_{\pm 0.34}$ & $91.04_{\pm 0.00}$
& $56.25_{\pm 0.08}$ & $55.49_{\pm 0.18}$
& $85.25$ & $77.94$ & $49.12$ & $75.35$
& 74.00
\\

\midrule

\multicolumn{10}{l}{\textbf{\textit{Qwen3-4B}}}
\\[-1pt]

ReAct
& $30.95_{\pm 1.47}$ & $24.38_{\pm 2.46}$
& $35.23_{\pm 0.56}$ & $30.49_{\pm 0.69}$
& $73.45_{\pm 0.28}$ & $73.87_{\pm 0.37}$
& $28.00_{\pm 3.53}$ & $68.90_{\pm 0.35}$
& 40.53
\\

GRPO
& $81.43_{\pm 1.54}$ & $71.89_{\pm 3.72}$
& $65.07_{\pm 0.14}$ & $49.69_{\pm 0.34}$
& $\bm{81.40_{\pm 0.45}}$ & $75.08_{\pm 0.09}$
& $\underline{36.33_{\pm 3.39}}$ & $\underline{76.49_{\pm 0.54}}$
& 67.12
\\

DAPO
& $\bm{91.67_{\pm 1.35}}$ & $\underline{86.57_{\pm 1.83}}$
& $\underline{66.08_{\pm 0.53}}$ & $\underline{51.48_{\pm 0.17}}$
& $77.62_{\pm 1.09}$ & $\underline{76.07_{\pm 0.33}}$ 
& $31.79_{\pm 3.84}$ & $67.88_{\pm 0.63}$
& \underline{70.41}
\\

GiGPO
& $\underline{87.38_{\pm 0.89}}$ & $84.58_{\pm 1.27}$
& $60.97_{\pm 0.44}$ & $49.59_{\pm 0.32}$
& $79.49_{\pm 0.70}$ & $75.67_{\pm 0.86}$ 
& $31.13_{\pm 2.91}$ & $74.81_{\pm 0.46}$ 
& 68.85
\\

HGPO
& $86.67_{\pm 0.89}$ & $82.84_{\pm 1.22}$
& $61.41_{\pm 0.50}$ & $49.56_{\pm 0.58}$
& $\underline{81.12_{\pm 0.40}}$ & $75.47_{\pm 0.40}$
& $33.25_{\pm 3.69}$ & $\bm{78.29_{\pm 0.36}}$
& 69.09
\\

\rowcolor{gray!15}
\textbf{\deltattg{} (Ours)}
& $\bm{91.67_{\pm 2.69}}$ & $\bm{87.81_{\pm 2.75}}$
& $\bm{67.79_{\pm 0.26}}$ & $\bm{52.88_{\pm 0.18}}$
& $80.29_{\pm 0.78}$ & $\bm{76.66_{\pm 1.03}}$ & $\bm{41.09_{\pm 2.40}}$ & $76.02_{\pm 0.60}$
& \textbf{72.86}
\\

\midrule

\multicolumn{10}{l}{\textbf{\textit{Llama-3.1-8B-Instruct}}}
\\[-1pt]

ReAct
& $16.90_{\pm 0.34}$ & $17.16_{\pm 1.22}$
& $26.96_{\pm 0.71}$ & $18.76_{\pm 0.77}$
& $46.33_{\pm 0.56}$ & $34.99_{\pm 0.75}$
& $7.62_{\pm 1.24}$ & $29.31_{\pm 0.99}$
& 23.15
\\

GRPO
& $77.38_{\pm 0.89}$ & $72.14_{\pm 1.96}$
& $66.15_{\pm 0.34}$ & $53.57_{\pm 0.42}$
& $70.48_{\pm 0.38}$ & $52.75_{\pm 0.85}$
& $10.21_{\pm 1.71}$ & $\underline{39.97_{\pm 1.95}}$
& 59.32
\\

DAPO
& $77.86_{\pm 1.54}$ & $74.13_{\pm 1.96}$
& $\underline{68.87_{\pm 0.28}}$ & $\underline{55.91_{\pm 0.28}}$
& $67.83_{\pm 1.55}$ & $45.20_{\pm 1.49}$
& $\underline{12.17_{\pm 2.70}}$ & $38.76_{\pm 0.39}$
& 59.79
\\

GiGPO
& $83.57_{\pm 1.01}$ & $\underline{82.34_{\pm 1.53}}$
& $65.19_{\pm 0.38}$ & $54.09_{\pm 0.58}$
& $69.15_{\pm 1.38}$ & $45.77_{\pm 0.77}$
& $12.08_{\pm 1.33}$ & $\bm{42.93_{\pm 0.20}}$
& \underline{61.69}
\\

HGPO
& $\underline{85.95_{\pm 0.34}}$ & $77.86_{\pm 1.96}$
& $64.73_{\pm 0.44}$ & $53.44_{\pm 0.48}$
& $\underline{71.24_{\pm 0.44}}$ & $\underline{56.55_{\pm 0.16}}$
& $10.54_{\pm 2.03}$ & $31.56_{\pm 1.39}$
& 61.15
\\

\rowcolor{gray!15}
\textbf{\deltattg{} (Ours)}
& $\bm{87.14_{\pm 0.58}}$ & $\bm{84.83_{\pm 0.70}}$
& $\bm{71.12_{\pm 0.54}}$ & $\bm{56.43_{\pm 0.19}}$
& $\bm{78.86_{\pm 0.53}}$ & $\bm{65.56_{\pm 0.24}}$ & $\bm{15.46_{\pm 2.42}}$ & $38.95_{\pm 0.22}$
& \textbf{66.49}
\\

\bottomrule

\end{tabular}
}

\end{table*}

\begin{figure*}[!t]
    \centering

    \begin{minipage}[t]{0.49\textwidth}
        \centering
        \includegraphics[width=\linewidth]
        {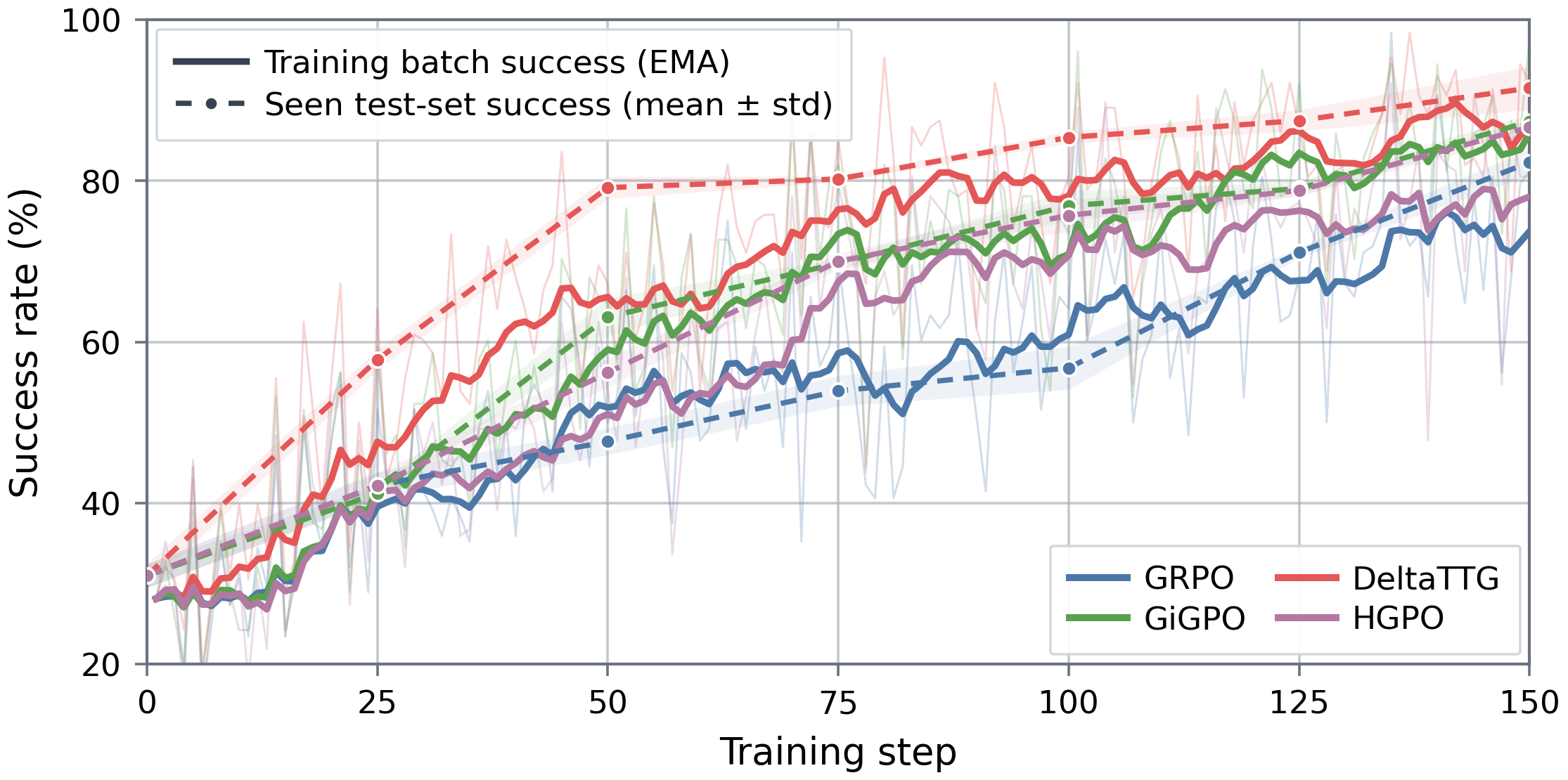}
        
    \end{minipage}
    \hfill
    \begin{minipage}[t]{0.49\textwidth}
        \centering
        \includegraphics[width=\linewidth]{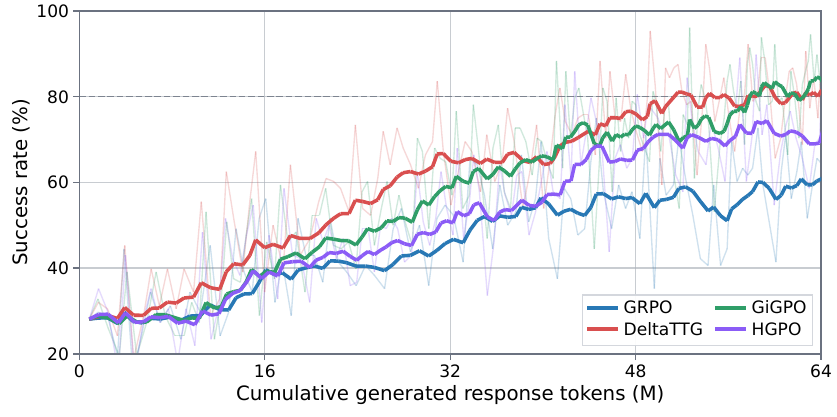}
        
    \end{minipage}

    \caption{
    Sample efficiency on ALFWorld Seen using Qwen3-4B.
    \textbf{(a)} Success rate as a function of training steps.
    \textbf{(b)} Training batch success rate as a function of cumulative actor-generated response tokens, excluding prompt and progress-meter tokens.
    Solid curves show the EMA-smoothed training success rate, while dashed curves and shaded regions in (a) denote the mean test success rate and standard deviation across three evaluation seeds.
    }
    \label{fig:alfworld_sample_efficiency}
\end{figure*}

\subsection{Main Results}

\paragraph{\deltattg{} achieves the best average performance with both backbones.}
Table~\ref{tab:main_results} shows that \deltattg{} achieves the highest average score with both backbones (72.86 vs.\ 70.41 for the strongest baseline with Qwen3-4B, and 66.49 vs.\ 61.69 with Llama-3.1-8B-Instruct) and the highest mean, including ties, on 13 of the 16 open-backbone metrics. The largest gains appear on BFCLv3 tool use: \deltattg{} improves Multi-Turn by 4.8 points over the strongest RL baseline with Qwen3-4B, and Non-Live and Live by 7.6 and 9.0 points with Llama-3.1-8B-Instruct. It trails the strongest baseline on Hallucination/Irrelevance with both backbones. These results span distinct interaction structures and support the effectiveness of progress-based credit allocation across different actor backbones and agent tasks.

\paragraph{\deltattg{} achieves better sample efficiency.}

Figure~\ref{fig:alfworld_sample_efficiency} compares the training dynamics of \deltattg{} and competing RL methods on ALFWorld Seen with Qwen3-4B. \deltattg{} reaches approximately 80\% test success by step 75, whereas GiGPO requires around 125 steps, corresponding to roughly 40\% fewer optimization steps in this training run. Panel (b) shows a similar advantage when progress is indexed by actor-generated response tokens. This axis excludes prompt and meter tokens, so it measures actor response-token efficiency rather than total training cost; Figure~\ref{fig:compute_budget} separately accounts for meter inference in GPU-hours. All training curves for the main experiments are provided in Appendix~\ref{app:detailed_training_curves}.

\subsection{Ablation Studies}

\begin{table}[t]
\centering
\caption{Ablation results on ALFWorld. $-$ denotes removal; parentheses show replacement signals.}
\label{tab:alfworld_ablation}
\scriptsize
\setlength{\tabcolsep}{2.2pt}
\renewcommand{\arraystretch}{1.05}
\resizebox{\linewidth}{!}{
\begin{tabular}{lccccccc}
\toprule
\textbf{Variant}
& \textbf{Pick}
& \textbf{Clean}
& \textbf{Heat}
& \textbf{Cool}
& \textbf{Look}
& \textbf{Two}
& \textbf{Avg.} \\
\midrule
\multicolumn{8}{l}{\textit{Seen}} \\
\midrule
\rowcolor{gray!15}
\textbf{Full \deltattg{}}
& 97.14 & 96.30 & 89.58 & 86.67
& 94.87 & 83.33 & \textbf{91.67} \\
$-$ TTG (+value)
& 97.14 & 85.19 & 81.25 & 76.00
& 76.92 & 87.50 & 85.71 \\
$-$ TTG (+logit)
& 98.10 & 92.59 & 95.83 & 78.67
& 89.74 & 63.89 & 86.67 \\
$-$ TTG (+feas.)
& 99.05 & 95.06 & 83.33 & 68.00
& 74.36 & 70.83 & 83.81 \\
$-$ sign-aware reversal
& 95.24 & 92.59 & 85.42 & 78.67
& 66.67 & 62.50 & 82.38 \\
$-$ meter training
& 99.05 & 62.96 & 79.17 & 81.33
& 76.92 & 80.56 & 81.43 \\
$-$ uniform mixing
& 96.19 & 100.00 & 87.50 & 78.67
& 82.05 & 84.72 & 89.52 \\
\midrule
\multicolumn{8}{l}{\textit{Unseen}} \\
\midrule
\rowcolor{gray!15}
\textbf{Full \deltattg{}}
& 87.50 & 93.55 & 84.06 & 85.71
& 90.74 & 82.35 & \textbf{87.81} \\
$-$ TTG (+value)
& 91.67 & 80.65 & 82.61 & 85.71
& 77.78 & 82.35 & 83.58 \\
$-$ TTG (+logit)
& 90.28 & 80.65 & 97.10 & 92.06
& 83.33 & 41.18 & 82.34 \\
$-$ TTG (+feas.)
& 86.11 & 92.47 & 86.96 & 85.71
& 94.44 & 62.75 & 85.82 \\
$-$ sign-aware reversal
& 75.00 & 77.42 & 86.96 & 80.95
& 85.19 & 50.98 & 76.87 \\
$-$ meter training
& 77.78 & 61.29 & 88.41 & 90.48
& 59.26 & 84.31 & 76.12 \\
$-$ uniform mixing
& 66.67 & 79.57 & 85.51 & 87.30
& 92.59 & 90.20 & 82.59 \\
\bottomrule
\end{tabular}
}
\end{table}

We consider six ablations on ALFWorld (Table~\ref{tab:alfworld_ablation}): (1) \textbf{$-$ TTG (+value)} trains a separate outcome-only meter on the same trajectories, using each trajectory's final score as the label for every prefix and keeping all other settings unchanged; (2) \textbf{$-$ TTG (+logit)} uses the clipped log-probability difference of the completion-label tokens; (3) \textbf{$-$ TTG (+feas.)} uses binary feasibility, mapping predicted successful and unsuccessful continuations to costs $0$ and $C$; (4) \textbf{$-$ sign-aware reversal} uses positive progress even on negative-advantage trajectories; (5) \textbf{$-$ meter training} uses the initialized progress meter without TTG-specific training; and (6) \textbf{$-$ uniform mixing} sets $\rho=0$, retaining equal shares only for the zero-mass fallback (definitions in Appendix~\ref{app:ablation_details}). The full \deltattg{} achieves higher average success than all ablations on both splits. The results favor TTG-based remaining-work supervision over outcome-value supervision here, support combining completion predictions with token estimates, and suggest the contributions of sign-aware reversal, meter training, and uniform mixing.

\subsection{Computational Cost}
\label{sec:computational_cost}

\begin{figure}[t]
\centering
\includegraphics[width=\columnwidth,trim=0 10bp 0 0,clip]{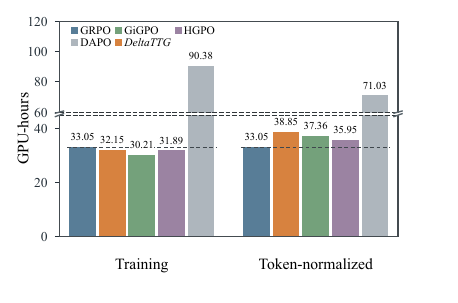}
\caption{Actor-training compute on ALFWorld, including progress-meter inference for \deltattg{}. Token-normalized GPU-hours use GRPO's 390.49M non-padding actor-training tokens, including prompts and responses, as the reference.}
\label{fig:compute_budget}
\end{figure}

Figure~\ref{fig:compute_budget} compares actor-training GPU-hours under a fixed 150-step ALFWorld budget. The measurement includes rollouts, actor forward passes and updates, and, for \deltattg{}, progress-meter inference; it excludes periodic evaluation and the separate progress-meter training stage. The frozen meter scores collected decision histories in batches and emits a short structured answer, limiting inference overhead relative to repeated actor generation and optimization. On ALFWorld Seen with Qwen3-4B, DAPO and \deltattg{} tie for the highest success rate. However, DAPO uses 2.7$\times$ GRPO's GPU-hours with dynamic sampling, while \deltattg{} achieves the same success rate at roughly GRPO's cost. The ALFWorld progress meter requires 10.45 GPU-hours of RL training and is reused across actor models, amortizing this cost across policy-training runs.

\subsection{Out-of-Domain Policy Generalization}
\label{sec:generalization}

\begin{figure*}[t]
    \centering

    \begin{subfigure}[t]{0.235\textwidth}
        \centering
        \includegraphics[width=\linewidth]{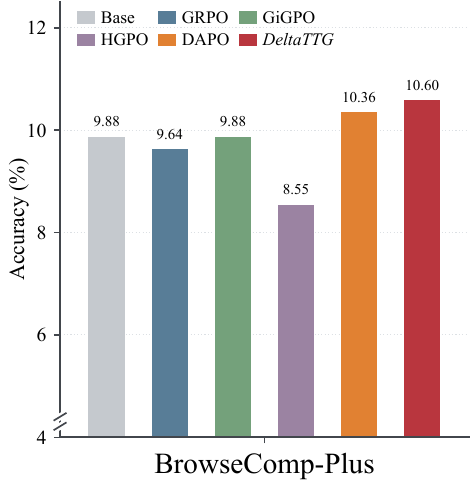}
        \caption{BrowseComp-Plus.}
        \label{fig:subfig_a}
    \end{subfigure}
    \hfill
    \begin{subfigure}[t]{0.235\textwidth}
        \centering
        \includegraphics[width=\linewidth]{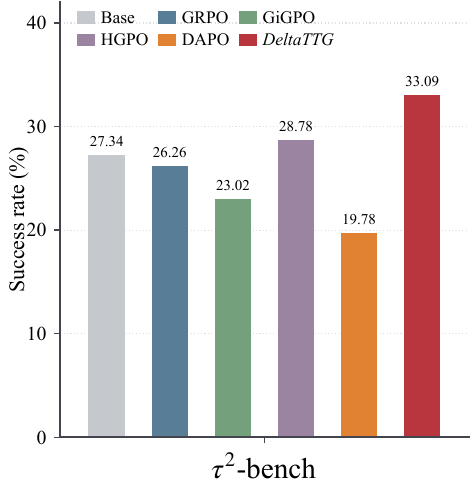}
        \caption{$\tau^2$-bench.}
        \label{fig:subfig_b}
    \end{subfigure}
    \hfill
    \begin{subfigure}[t]{0.235\textwidth}
        \centering
        \includegraphics[width=\linewidth]{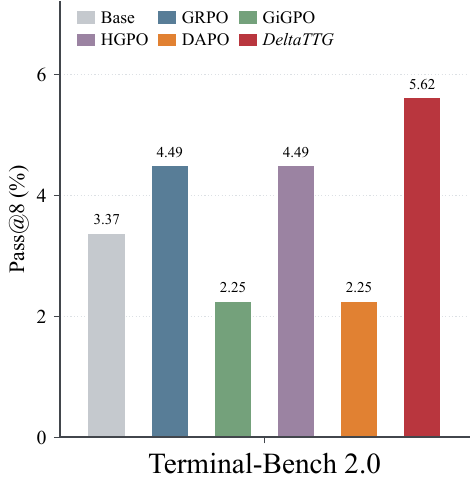}
        \caption{Terminal-Bench 2.0.}
        \label{fig:subfig_c}
    \end{subfigure}
    \hfill
    \begin{subfigure}[t]{0.235\textwidth}
        \centering
        \includegraphics[width=\linewidth]{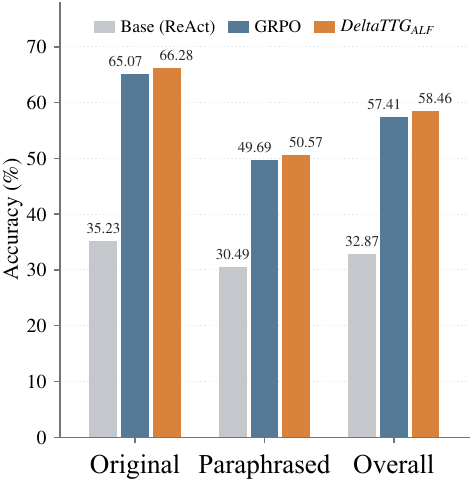}
        \caption{Progress-meter transfer.}
        \label{fig:subfig_d}
    \end{subfigure}

    \caption{
    \textbf{Generalization analysis of \deltattg{}.}
    \textbf{(a)--(c)} Out-of-domain generalization of Qwen3-4B policies trained on PaperSearchQA for (a) and AgentWorldModel for (b)--(c). Panel (a) uses a truncated y-axis starting at 4\%, with an axis-break mark; panel (c) reports pass@8 on 89 tasks.
    \textbf{(d)} Transfer of an ALFWorld-trained progress meter to \deltattg{} optimization on PaperSearchQA, assessing progress signals beyond the meter's training domain.
    }
    \label{fig:deltattg_generalization}
\end{figure*}

\paragraph{\deltattg{} improves in-domain performance while better preserving out-of-domain capabilities.}
We evaluate Qwen3-4B checkpoints on the three out-of-domain benchmarks of Section~\ref{sec:experimental_setup}. For the search task BrowseComp-Plus, we use checkpoints trained on PaperSearchQA; for the complex engineering tasks $\tau^2$-bench and Terminal-Bench 2.0, we use checkpoints trained on AgentWorldModel. \deltattg{} achieves the highest performance among the compared methods on all three benchmarks (Figures~\ref{fig:subfig_a}--\ref{fig:subfig_c}; $\tau^2$-bench domains in Appendix~\ref{app:ood_results}). On these benchmarks, the results suggest that task-progress-based credit assignment improves in-domain task performance while preserving capabilities on the evaluated out-of-domain benchmarks. Detailed out-of-domain evaluation settings are provided in Appendix~\ref{app:ood_settings}.

\section{Analyzing the Progress Meter}
\label{sec:analysis}
We now evaluate the progress meter itself through three questions: Is Token-to-Go a good supervision target for step quality (Section~\ref{sec:ttg_supervision})? Does a fixed meter help actors select actions at test time (Section~\ref{sec:test_time_guidance})? Does a meter transfer across domains (Section~\ref{sec:progress_meter_transfer})?

\subsection{Token-to-Go as Progress Supervision}
\label{sec:ttg_supervision}

\paragraph{Token-to-Go supervision provides a step-quality signal.}
We train a TTG progress meter and use AgentProcessBench's step-quality annotations~\citep{fan2026agentprocessbench} to evaluate whether changes in its estimates across consecutive states distinguish beneficial from harmful actions (Table~\ref{tab:apb_supervision}). With neutral steps excluded, uniform random guessing has an expected accuracy of 50\%; direct prompting achieves 50.83\% on the held-out test split, compared with 56.91\% for Token-to-Go supervision. This analysis jointly trains a meter on four datasets, whereas actor-training experiments use domain-specific meters (Appendix~\ref{app:apb_analysis}).

\paragraph{Token-to-Go achieves stronger overall accuracy than alternative supervision targets.}
We compare Token-to-Go with Future Score and Step-to-Go, which supervise terminal returns and remaining-step counts, respectively. Token-to-Go achieves the highest step-pooled Overall accuracy on both validation and test splits (Table~\ref{tab:apb_supervision}). TTG labels vary across trajectory prefixes and reflect remaining workload, potentially providing more informative local supervision than terminal-return labels, which can lead to \emph{value flattening} under sparse rewards~\citep{li2026rethinking}. The ALFWorld ablations provide further downstream evidence: replacing Token-to-Go with completion logits or binary feasibility reduces Unseen success (Table~\ref{tab:alfworld_ablation}).

\begin{table}[t]
\centering
\small
\setlength{\tabcolsep}{2pt}
\begin{tabular}{@{}lrrrrr@{}}
\toprule
\makecell[l]{Supervision\\target} & BFCL & GAIA & \makecell{Hotpot\\QA} & \makecell{$\tau^2$-\\bench} & Overall \\
\midrule
\multicolumn{6}{l}{\textit{Validation}} \\
Future Score & 53.04 & 70.89 & 44.12 & 63.89 & 60.22 \\
Step-to-Go & 57.39 & 59.49 & 52.94 & 53.47 & 55.91 \\
\rowcolor{gray!15}
Token-to-Go & 65.22 & 60.76 & 52.94 & 61.81 & \textbf{61.83} \\
\midrule
\multicolumn{6}{l}{\textit{Test}} \\
Future Score & 55.47 & 60.96 & 47.66 & 53.57 & 54.86 \\
Step-to-Go & 61.52 & 58.57 & 61.72 & 47.27 & 54.73 \\
\rowcolor{gray!15}
Token-to-Go & 54.49 & 52.19 & 53.91 & 60.84 & \textbf{56.91} \\
\bottomrule
\end{tabular}
\caption{AgentProcessBench validation and test step-quality accuracy (\%). Overall pools evaluated steps across datasets; the best Overall in each split is bold. See Appendix~\ref{app:apb_analysis} for the evaluation protocol.}
\label{tab:apb_supervision}
\end{table}

\begin{figure}[t]
\centering
\includegraphics[width=\linewidth]{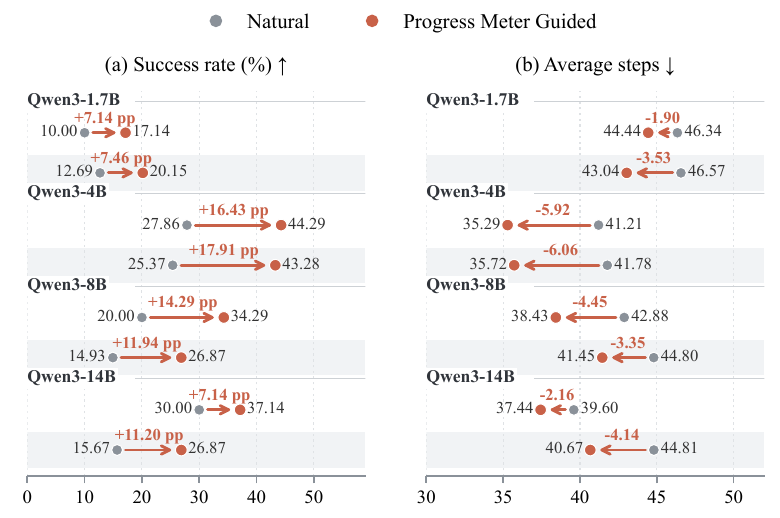}
\caption{ALFWorld test-time guidance results. Arrows connect Natural to Progress Meter Guided rollouts; labels show changes in success rate (percentage points) and average steps. White/gray rows denote Seen/Unseen. The protocol is in Appendix~\ref{app:test_time_progress}.}
\label{fig:test_time_progress}
\end{figure}

\subsection{Test-Time Guidance across Actor Scales}
\label{sec:test_time_guidance}

\paragraph{Token-to-Go progress meters improve test-time action selection across actor scales.}
We evaluate whether a shared progress meter can provide useful progress estimates during inference. On ALFWorld, a fixed meter guides Qwen3-1.7B, 4B, 8B, and 14B actors without actor-specific retraining. At each step, the actor generates four candidate actions, which are trial-executed in separate environment branches; the meter evaluates the resulting histories to select the action with the largest estimated progress. Compared with natural rollouts and random selection using the same four-candidate budget as guidance, guided rollouts achieve higher success rates and fewer average interaction steps across all evaluated actor scales and Seen/Unseen splits (Figure~\ref{fig:test_time_progress}; Appendix~\ref{app:test_time_progress}, Table~\ref{tab:test_time_three_way}). The comparison with random selection indicates that the gains arise from progress-based action selection rather than additional candidates alone.

\subsection{Cross-Domain Progress-Meter Transfer}
\label{sec:progress_meter_transfer}

\paragraph{\normalfont\bfseries Progress meters provide useful progress signals beyond their training domain.}
\deltattgvar{ALF} uses a progress meter trained only on ALFWorld trajectories to guide actor optimization on PaperSearchQA, without target-domain adaptation. It outperforms GRPO on Original, Paraphrased, and Overall evaluation (Figure~\ref{fig:subfig_d}), suggesting that learned progress estimates can support credit assignment beyond the training domain.

\section{Conclusion}
We introduced \deltattg{}, which uses a reusable Token-to-Go meter trained on recorded trajectories to redistribute outcome-derived advantages. It improves over strong RL baselines at actor-training cost close to GRPO and supports test-time action selection and cross-domain meter transfer.

\section*{Limitations}
Our study has three limitations. First, although diverse training trajectories support progress-meter reuse, our evaluation covers only two actor model families and several domains; broader transfer requires further validation and may need adaptation. Second, our AgentWorldModel audit (Section~\ref{sec:deltattg}, footnote~\ref{fn:failure_cost_audit}) indicates that $C$ does not affect credit allocation for most successful trajectories, and using task-specific values across three training domains yields higher aggregate scores than all compared RL baselines. However, these observations do not establish robustness to $C$, and future work will provide a more detailed sensitivity analysis. Third, reported standard deviations reflect repeated evaluations of a single trained checkpoint rather than independent training runs, leaving training variability unquantified.

\bibliography{iclr2027_conference}

\appendix
\section{Benchmark}
\label{app:benchmarks}

\paragraph{ALFWorld.}
ALFWorld~\citep{shridhar2020alfworld} is a text-based household benchmark in which agents navigate rooms and manipulate objects to satisfy task goals. We use 3,553 tasks for training and evaluate on the Seen (140 tasks) and Unseen (134 tasks) splits. Both evaluation splits cover six task types: Pick \& Place (35/24), Clean \& Place (27/31), Heat \& Place (16/23), Cool \& Place (25/21), Examine in Light (13/18), and Pick Two \& Place (24/17), where counts are given as Seen/Unseen.

\paragraph{PaperSearchQA.}
PaperSearchQA~\citep{burgess2026papersearchqa} is a biomedical question-answering benchmark that requires agents to retrieve evidence from 16 million PubMed abstracts. We use 54,907 training examples, comprising 27,396 Original and 27,511 Paraphrased questions, and 5,000 test examples, with 2,500 in each subset. We report exact-match success separately on the Original and Paraphrased test subsets.

\paragraph{AgentWorldModel.}
AgentWorldModel (AWM)~\citep{wang2026agentworldmodel} provides synthetic, executable tool-use environments with database-backed state transitions for multi-turn agent interaction. Our AWM data comprise 9,850 tasks across 985 environments, all used for actor training; AWM is not used as a test benchmark.

\paragraph{BFCLv3.}
BFCLv3~\citep{patil2025bfcl} evaluates function selection, argument generation, and multi-turn tool use. We use it only for testing and report four categories: Non-Live (1,150 cases), Live (1,351), Multi-Turn (800), and Hallucination/Irrelevance (1,122). Multi-Turn contains Base, Missing Function, Missing Parameter, and Long Context subsets (200 each); Hallucination/Irrelevance combines 240 non-live and 882 live cases requiring the agent to refrain from an unsupported function call.

\paragraph{BrowseComp-Plus.}
BrowseComp-Plus~\citep{chen2025browsecompplus} evaluates difficult information-seeking questions through iterative retrieval over a fixed corpus of 100,195 documents with verified supporting evidence. We use all 830 questions solely for out-of-domain testing and report answer accuracy over the full evaluation set.

\paragraph{$\tau^2$-bench.}
$\tau^2$-bench~\citep{barres2025tau2bench} evaluates conversational agents that coordinate with users and invoke tools to complete service tasks. We use 278 tasks solely for out-of-domain testing, comprising Airline (50), Retail (114), and Telecom (114), and report overall and domain-level success rates.

\paragraph{Terminal-Bench 2.0.}
Terminal-Bench 2.0~\citep{merrill2026terminalbench} evaluates agents on realistic command-line tasks, including software development, system administration, and data processing, with executable outcome checks. We use all 89 tasks solely for out-of-domain testing and report pass@8 over this evaluation set.

\section{Baseline Methods}
\label{app:baselines}
\label{sec:baseline_methods}

We compare DeltaTTG with a diverse set of baseline methods, including
prompt-based agent methods and reinforcement learning approaches.
The prompt-based baselines evaluate the capability of the pretrained
models without parameter updates, while the RL baselines represent
different strategies for optimizing agent policies from environment
feedback.

\paragraph{ReAct.}
ReAct~\citep{yao2022react} interleaves reasoning and executable actions using environment observations and recent interaction history. We retain the task interface and per-attempt interaction limits of the corresponding RL evaluation.

\paragraph{Reflexion.}
Reflexion~\citep{shinn2023reflexion} retries failed tasks with textual reflections in its context. The ALFWorld and PaperSearchQA configurations allow up to 10 attempts and retain the three most recent reflections. BFCLv3 uses up to three attempts, with at most 512 generated tokens per reflection. Success is counted if any attempt within the budget succeeds; completed tasks are not retried.

\paragraph{GRPO.}
GRPO~\citep{shao2024deepseekmath} assigns each trajectory a group-relative outcome advantage without training a critic. We use the vanilla clipped policy loss with token-mean reduction and one PPO epoch per update. The implementation uses dual clipping with coefficient 3 for negative advantages.

\paragraph{DAPO.}
DAPO~\citep{yu2026dapo} uses asymmetric clipping with lower/upper widths 0.2/0.28, token-mean reduction, and dynamic sampling. We disable the actor KL loss and filter groups with reward standard deviation below $10^{-4}$, allowing up to 10 generation batches per update. The task interface and backbone follow GRPO, while dynamic sampling can require additional generated trajectories to form an accepted training batch.

\paragraph{GiGPO.}
GiGPO~\citep{feng2026group} supplements trajectory-level credit with advantages computed among actions at matched states. We use discount factor $\gamma=0.95$, step-advantage weight 1, and mean/std normalization. ALFWorld uses exact state matching; PaperSearchQA and AgentWorldModel enable similarity matching with threshold 0.95. Step-group sizes are determined by the matched state visits rather than a fixed group-size hyperparameter.

\paragraph{HGPO.}
HGPO~\citep{he2026hierarchy} groups steps by progressively longer matching histories and combines the resulting relative advantages. We use $\gamma=0.95$, mean/std normalization, and history-length weighting with exponent 1 (\texttt{weight\_type=length}, \texttt{base\_group=False}).

\paragraph{GPT-5.6-Sol.}
We evaluate GPT-5.6-Sol as a proprietary-model baseline with reasoning effort set to \texttt{low}.

\paragraph{Claude-Opus-5.}
We evaluate Claude-Opus-5 as a proprietary-model baseline with reasoning effort set to \texttt{low}.

\section{Implementation Details}
\label{app:implementation}

\subsection{Training Infrastructure}

For all agentic RL experiments, we build upon
\texttt{verl-agent}\footnote{\url{https://github.com/langfengQ/verl-agent}},
an extension of \texttt{veRL}~\citep{sheng2024hybridflow} designed for
reinforcement learning of LLM agents.
We use its rollout and distributed training infrastructure as the basis for
all RL methods, and integrate DeltaTTG into the same training pipeline to
ensure consistent rollout and optimization settings across different methods.

\subsection{Computational Resources}

The AgentProcessBench analysis in Section~5.1 is conducted using AMD Instinct MI325X GPUs. All other experiments use NVIDIA H200 GPUs.

\subsection{Training Hyperparameters}
\label{sec:predictor_details}

Table~\ref{tab:hyperparams} summarizes the repository defaults for common training hyperparameters and method-specific settings. A dash denotes a setting that does not apply.

\begin{table*}[t]
\centering
\small
\setlength{\tabcolsep}{3pt}
\renewcommand{\arraystretch}{1.1}
\begin{tabular*}{\textwidth}{@{\extracolsep{\fill}}lcccccc@{}}
\toprule
\multirow{2}{*}{\textbf{Hyperparameter}} & \multicolumn{2}{c}{\textbf{ALFWorld}} & \multicolumn{2}{c}{\textbf{PaperSearchQA}} & \multicolumn{2}{c}{\textbf{AgentWorldModel}} \\
\cmidrule(lr){2-3}\cmidrule(lr){4-5}\cmidrule(lr){6-7}
 & Qwen & Llama & Qwen & Llama & Qwen & Llama \\
\midrule
\multicolumn{7}{@{}l}{\textbf{Common settings}} \\
Learning rate & $10^{-6}$ & $10^{-6}$ & $10^{-6}$ & $10^{-6}$ & $10^{-6}$ & $10^{-6}$ \\
Training batch size & 16 & 16 & 512 & 512 & 16 & 16 \\
Rollout group size $G$ & 8 & 8 & 5 & 5 & 16 & 16 \\
PPO epochs & 1 & 1 & 1 & 1 & 1 & 1 \\
Max turns & 50 & 50 & 6 & 6 & 20 & 20 \\
Max tokens per turn & 1,024 & 1,024 & 1,024 & 1,024 & 2,048 & 2,048 \\
Temperature & 1.0 & 1.0 & 1.0 & 1.0 & 1.0 & 0.8 \\
Top-$p$ & 1.0 & 1.0 & 1.0 & 1.0 & 1.0 & 1.0 \\
KL coefficient & 0.01 & 0.01 & 0.001 & 0.001 & 0.001 & 0.01 \\
Entropy coefficient & 0.001 & 0.001 & 0 & 0 & 0 & 0 \\
Clip lower / upper & 0.2 / 0.28 & 0.2 / 0.28 & 0.2 / 0.28 & 0.2 / 0.28 & 0.2 / 0.28 & 0.2 / 0.28 \\
Dual clip & 3 & 3 & 3 & 3 & 3 & 3 \\
\midrule
\multicolumn{7}{@{}l}{\textbf{Method-specific settings}} \\
\rowcolor{gray!15}
\multicolumn{7}{@{}l}{\textit{DAPO}} \\
Clip lower / upper & 0.2 / 0.28 & 0.2 / 0.28 & 0.2 / 0.28 & 0.2 / 0.28 & 0.2 / 0.28 & 0.2 / 0.28 \\
Max sampling batches & 10 & 10 & 10 & 10 & 10 & 10 \\
Min reward std. & $10^{-4}$ & $10^{-4}$ & $10^{-4}$ & $10^{-4}$ & $10^{-4}$ & $10^{-4}$ \\
\rowcolor{gray!15}
\multicolumn{7}{@{}l}{\textit{GiGPO/HGPO}} \\
Step weight & 1 & 1 & 1 & 1 & 1 & 1 \\
State matching & Exact & Exact & Similarity & Similarity & Similarity & Similarity \\
Similarity threshold & -- & -- & 0.95 & 0.95 & 0.95 & 0.95 \\
History exponent & 1 & 1 & 1 & 1 & 1 & 1 \\
\rowcolor{gray!15}
\multicolumn{7}{@{}l}{\textit{DeltaTTG}} \\
Mixing ratio $\rho$ & 0.2 & 0.2 & 0.2 & 0.2 & 0.2 & 0.2 \\
Multiplier cap $M$ & 5 & 5 & 5 & 5 & 5 & 5 \\
Failure cost $C$ & 16,384 & 16,384 & 6,144 & 6,144 & 16,384 & 16,384 \\
\bottomrule
\end{tabular*}
\caption{Training hyperparameters by dataset and actor backbone. Qwen and Llama denote Qwen3-4B and Llama-3.1-8B-Instruct.}
\label{tab:hyperparams}
\end{table*}

\paragraph{Progress-meter hyperparameters.}
The ALFWorld and PaperSearchQA repository configurations use an SFT learning rate of $10^{-5}$ for one epoch, followed by GRPO refinement with a learning rate of $3\times10^{-6}$ and a group size of 8. Each domain's meter is shared by both actor backbones. The training procedure is described in Appendix~\ref{app:progress_meter_training}.

\subsection{Evaluation Metrics}
\label{sec:evaluation_metrics}

We report success rates on ALFWorld (the task goal is satisfied) and PaperSearchQA (exact match against any accepted answer), and the official category scores on BFCLv3. For repeated evaluations in Table~\ref{tab:main_results}, we report the mean and population standard deviation over three test runs; entries without a standard deviation report the available point estimate.

\subsection{Out-of-Domain Evaluation Settings}
\label{app:ood_settings}

All three evaluations use Qwen3-4B actors. BrowseComp-Plus uses checkpoints trained on PaperSearchQA, while Terminal-Bench 2.0 and $\tau^2$-bench use checkpoints trained on AgentWorldModel. Table~\ref{tab:ood_decoding} summarizes actor decoding and serving settings. Thinking is enabled via \texttt{enable\_thinking=True}; for Terminal-Bench 2.0, this setting refers to the main agent. For $\tau^2$-bench, the user simulator uses GPT-5.2 with reasoning effort set to \texttt{low}; all three domains (Airline, Retail, and Telecom) use non-solo mode.

\begin{table}[t]
\centering
\footnotesize
\setlength{\tabcolsep}{2pt}
\renewcommand{\arraystretch}{1.15}
\begin{tabular*}{\columnwidth}{@{\extracolsep{\fill}}lccc@{}}
\toprule
Parameter & \makecell{Terminal-\\Bench 2.0} & $\tau^2$-bench & \makecell{BrowseComp-\\Plus} \\
\midrule
Temperature & 0.6 & 0.6 & 0.7 \\
Top-$p$ & 0.95 & 0.95 & 0.8 \\
Top-$k$ & 20 & 20 & 20 \\
Thinking & Enabled & Enabled & Enabled \\
Max output tokens & 8,192 & 2,048 & 10,000 \\
Context limit & 32,768 & 32,768 & 40,960 \\
\bottomrule
\end{tabular*}
\caption{Actor decoding and serving settings for out-of-domain evaluation. Output limits are per response; context limits are in tokens.}
\label{tab:ood_decoding}
\end{table}

\section{Method Details}
\label{app:method_details}

\subsection{Progress-Meter Training Data and Reuse}
\label{app:meter_data_reuse}

We collect raw agent trajectories on the training sets of ALFWorld, PaperSearchQA, and AgentWorldModel using a range of source models, including Qwen3-4B-Instruct-2507, Qwen3-8B, Qwen3.5-9B, Qwen3.5-27B, and Llama-3.2-3B-Instruct, among others. We also incorporate reference answers provided in the training sets of some datasets. We construct trajectory-prefix supervision from these records and train a separate progress meter for each task category, as summarized in Table~\ref{tab:meter_data_reuse}. In the main experiments, each domain's meter is frozen and shared by the Qwen3-4B and Llama-3.1-8B-Instruct actors during RL; it is not retrained separately for each actor backbone.

\begin{table}[htbp]
\centering
\small
\setlength{\tabcolsep}{3pt}
\begin{tabular}{@{}p{0.43\columnwidth}p{0.53\columnwidth}@{}}
\toprule
Trajectory source & Progress-meter task category \\
\midrule
ALFWorld (train) & Text-based interaction \\
PaperSearchQA (train) & Information seeking \\
AgentWorldModel (train) & Complex engineering tasks \\
\bottomrule
\end{tabular}
\caption{Training domains of the three progress meters. Each meter is reused for both actor backbones in the corresponding main experiments.}
\label{tab:meter_data_reuse}
\end{table}

\subsection{Token-to-Go Progress Meter Training}
\label{app:progress_meter_training}

\paragraph{Supervised fine-tuning.}
Each example pairs an observed prefix $h$ with a structured answer $y$ containing the completion label $b$ and remaining-token label $v$. For successful trajectories, $v$ is the recorded suffix token count; otherwise it is null. We use the answer-token cross-entropy in Equation~\ref{eq:meter_sft}, masking prefix and prompt tokens from the loss. This stage teaches the output format and initializes prediction from trajectory supervision.

\paragraph{Reinforcement learning.}
We refine the meter with GRPO using the reward and token-accuracy score in Equations~\ref{eq:meter_reward}--\ref{eq:meter_token_accuracy}.
Outputs are parsed before outcome and token-count scoring. For $G_{\mathrm{pred}}$ answers sampled for the same prefix, predictor RL normalizes rewards by their group mean $\bar r^{\mathrm{pred}}$ and sample standard deviation $s^{\mathrm{pred}}$:
\begin{equation}
A_j^{\mathrm{pred}}=\frac{r_j^{\mathrm{pred}}-\bar r^{\mathrm{pred}}}{s^{\mathrm{pred}}+\epsilon},\qquad \epsilon=10^{-6}.
\end{equation}
These rewards train only the progress meter; they are never added to the actor's environment reward. The meter is frozen throughout actor optimization.

\subsection{Group-Relative Advantage Normalization}
\label{app:group_advantages}

\label{sec:group_advantages}
Our implementation computes reward statistics over step-expanded training rows, following the default episode-level normalization in the GiGPO implementation~\citep{feng2026group}. Trajectory $i$ contributes $T_i$ rows carrying reward $R_i$. With $K_x=\sum_iT_i$, the mean, sample standard deviation, and advantage are
\begin{equation}
\begin{aligned}
\mu_x&=\frac{1}{K_x}\sum_{i=1}^{G}T_iR_i,\\
\sigma_x&=\sqrt{\frac{1}{K_x-1}\sum_{i=1}^{G}T_i(R_i-\mu_x)^2},\\
A_i&=\frac{R_i-\mu_x}{\sigma_x+\epsilon},\qquad\epsilon=10^{-6}.
\end{aligned}
\end{equation}
Thus, trajectories contribute to the group statistics in proportion to their numbers of actions. All trainable tokens initially receive the same $A_i$ before progress-based redistribution.

\subsection{Bounded Token-Credit Normalization}
\label{app:bounded_normalization}
To compute $\mathcal B_M$ in Equation~\ref{eq:token_credit_weights}, we broadcast $q_{i,t}$ to all trainable tokens of action $t$ and normalize to token mean one, giving $N_iq_{i,t}/\sum_s n_{i,s}q_{i,s}$ before capping each multiplier at $M$. The weight removed by capping is distributed among tokens below the cap in proportion to their remaining capacity $M-w$. This preserves unit token mean while keeping every final multiplier at most $M$.

\subsection{Actor Optimization Objective}
\label{app:actor_objective}
Let $z_{i,t,k}$ be the full token prefix preceding the $k$-th trainable token $a_{i,t,k}$ in action $t$. The importance ratio and its clipped version are
\begin{equation}
\begin{aligned}
r_{i,t,k}(\theta)
&=\frac{\pi_\theta(a_{i,t,k}\mid z_{i,t,k})}
{\pi_{\theta_{\mathrm{old}}}(a_{i,t,k}\mid z_{i,t,k})},\\
\bar r_{i,t,k}(\theta)
&=\operatorname{clip}(r_{i,t,k}(\theta),1-\eta_{\mathrm{low}},1+\eta_{\mathrm{high}}).
\end{aligned}
\end{equation}
We use $\eta_{\mathrm{low}}=0.2$ and $\eta_{\mathrm{high}}=0.28$, giving a clipping interval of $[0.8,1.28]$.
With $\widetilde A_{i,t,k}=A_iw_{i,t}$ fixed during the actor update, the token surrogate is
\begin{equation}
s_{i,t,k}(\theta)=\min\!\left(
 r_{i,t,k}(\theta)\widetilde A_{i,t,k}\!,
 \bar r_{i,t,k}(\theta)\widetilde A_{i,t,k}\right).
\end{equation}
The implementation additionally applies dual clipping with $d=3$ to negative advantages:
\begin{equation}
s^{\mathrm{dual}}_{i,t,k}(\theta)=
\begin{cases}
\max(s_{i,t,k}(\theta),d\widetilde A_{i,t,k}),&\widetilde A_{i,t,k}<0,\\
s_{i,t,k}(\theta),&\widetilde A_{i,t,k}\geq0.
\end{cases}
\end{equation}
The policy-gradient loss uses token-mean reduction,
\begin{equation}
\mathcal L_{\mathrm{PG}}(\theta)
=-\frac{1}{\sum_iN_i}\sum_{i,t,k}s^{\mathrm{dual}}_{i,t,k}(\theta).
\end{equation}
Configured KL regularization and entropy terms are added by the training framework; their settings follow the corresponding actor configuration.

\subsection{Algorithm}

Algorithm~\ref{alg:deltattg} presents the training pipeline of DeltaTTG.
After collecting complete actor trajectories and computing the
group-relative advantages from step-expanded statistics, we batch-score decision states of nonzero-advantage trajectories with a frozen TTG progress meter in a separate post-rollout stage.
The resulting TTG differences are then used to reweight the original advantages over actions and tokens, before the actor update. The policy loss and KL configuration are kept fixed when comparing the corresponding baseline with DeltaTTG.

\begin{algorithm}[t]
\caption{DeltaTTG Actor Training}
\label{alg:deltattg}
\footnotesize
\definecolor{algrollout}{HTML}{356B9A}
\definecolor{algmeter}{HTML}{27806A}
\definecolor{algcredit}{HTML}{976123}
\definecolor{algupdate}{HTML}{78569B}
\begin{algorithmic}[1]
\Require Actor $\pi_\theta$; frozen progress meter $f_\phi$; task distribution $p(\mathcal X)$; group size $G$; mixing ratio $\rho$; failure cost $C>0$; multiplier cap $M\geq1$
\For{each training iteration}
    \Statex \textcolor{algrollout}{\textit{// Collect trajectories and outcome advantages}}
    \State $\theta_{\mathrm{old}}\leftarrow\theta$; sample $x\sim p(\mathcal X)$
    \State Sample $G$ trajectories with $\pi_{\theta_{\mathrm{old}}}$ and observe terminal rewards $\{R_i\}$
    \State Compute $\{A_i\}$ from step-expanded group statistics
    \State Initialize $\widetilde A_{i,t,k}\leftarrow A_i$ for all trainable tokens
    \Statex \textcolor{algmeter}{\textit{// Evaluate states with the frozen progress meter}}
    \State Extract and deduplicate decision prefixes from trajectories with $A_i\ne0$
    \State Batch-query $f_\phi$ to obtain $\{(\hat b_{i,t},\hat v_{i,t})\}$
    \Statex \textcolor{algcredit}{\textit{// Redistribute credit when meter outputs are valid}}
    \If{all queried outputs for this prompt are valid}
        \For{each trajectory with $A_i\ne0$}
            \If{all queried states indicate failure}
                \State \textbf{continue} \Comment{Keep original advantages}
            \EndIf
            \State $c_{i,t}\leftarrow\hat v_{i,t}$ if $\hat b_{i,t}=1$; else $C$
            \State $c_{i,T_i}\leftarrow0$ for terminal success; else $C$
            \State $\Delta_{i,t}\leftarrow c_{i,t}-c_{i,t+1}$
            \State $m_{i,t}\leftarrow\max(\operatorname{sign}(A_i)\Delta_{i,t},0)$
            \If{$\sum_t m_{i,t}=0$}
                \State $q_{i,t}\leftarrow1/T_i$
            \Else
                \State $q_{i,t}\leftarrow\rho/T_i+(1-\rho)m_{i,t}/\sum_s m_{i,s}$
            \EndIf
            \State $w^{(0)}_{i,t}\leftarrow q_{i,t}$
            \State $\{w_{i,t}\}_t\leftarrow\mathcal B_M(\{w^{(0)}_{i,t}\}_t;\{n_{i,t}\}_t)$
            \State $\widetilde A_{i,t,k}\leftarrow A_iw_{i,t}$
        \EndFor
    \Else
        \State Keep original advantages for the entire prompt group
    \EndIf
    \Statex \textcolor{algupdate}{\textit{// Update the actor using redistributed advantages}}
    \State Update $\pi_\theta$ with the vanilla token-mean clipped loss and configured KL term
\EndFor
\end{algorithmic}
\end{algorithm}

\section{Additional Results}
\label{app:additional_results}

\subsection{Detailed Results on ALFWorld}

To provide a more fine-grained evaluation, we further report the performance
of all methods on the six ALFWorld task categories for both the \textit{Seen}
and \textit{Unseen} splits in Table~\ref{tab:alfworld_detailed}.
This breakdown allows us to examine whether the improvements of DeltaTTG are
consistent across different types of household manipulation tasks, rather than
being driven by a small subset of task categories.

\begin{table*}[t]
\centering
\caption{
Detailed performance comparison across different ALFWorld task types.
We report success rates ($\uparrow$) on the \textit{Seen} and
\textit{Unseen} splits. The \#Tasks row denotes the number of
evaluation instances in each category. Overall is the task-count-weighted average of the category success rates.
}
\label{tab:alfworld_detailed}

\begin{subtable}[t]{0.49\textwidth}
\centering
\caption{Qwen3-4B}
\label{tab:alfworld_detailed_qwen}

\scriptsize
\resizebox{\linewidth}{!}{
\begin{tabular}{lccccccc}
\toprule
\textbf{Method}
& \textbf{Pick}
& \textbf{Clean}
& \textbf{Heat}
& \textbf{Cool}
& \textbf{Look}
& \textbf{Two-Obj}
& \textbf{Overall} \\
\midrule

\multicolumn{8}{l}{\textit{Seen}} \\
\#Tasks
& 35 & 27 & 16 & 25 & 13 & 24 & 140 \\
\midrule

ReAct
& 62.86 & 19.75 & 33.33 & 9.33
& 38.46 & 13.89 & 30.95 \\

Reflexion
& 80.00 & 48.15 & 50.00 & 29.33
& 61.54 & 38.89 & 52.62 \\

\midrule

GRPO
& 92.38 & 87.65 & 79.17 & 77.33
& 71.79 & 69.44 & 81.43 \\

DAPO
& 98.10 & 98.77 & 89.58 & 84.00
& 84.62 & 87.50 & 91.67 \\

GiGPO
& 97.14 & 92.59 & 72.92 & 82.67
& 84.62 & 83.33 & 87.38 \\

HGPO
& 95.24 & 93.83 & 85.42 & 84.00
& 89.74 & 68.06 & 86.67 \\

\textbf{DeltaTTG}
& 97.14 & 96.30 & 89.58 & 86.67
& 94.87 & 83.33 & 91.67 \\

\midrule

\multicolumn{8}{l}{\textit{Unseen}} \\
\#Tasks
& 24 & 31 & 23 & 21 & 18 & 17 & 134 \\
\midrule

ReAct
& 51.39 & 18.28 & 20.29 & 23.81
& 27.78 & 0.00 & 24.38 \\

Reflexion
& 73.61 & 34.41 & 52.17 & 47.62
& 48.15 & 29.41 & 47.76 \\

\midrule

GRPO
& 88.89 & 72.04 & 73.91 & 84.13
& 57.41 & 45.10 & 71.89 \\

DAPO
& 81.94 & 79.57 & 84.06 & 90.48
& 100.00 & 90.20 & 86.57 \\

GiGPO
& 81.94 & 84.95 & 84.06 & 92.06
& 72.22 & 92.16 & 84.58 \\

HGPO
& 80.56 & 80.65 & 85.51 & 85.71
& 96.30 & 68.63 & 82.84 \\

\textbf{DeltaTTG}
& 87.50 & 93.55 & 84.06 & 85.71
& 90.74 & 82.35 & 87.81 \\

\bottomrule
\end{tabular}
}
\end{subtable}
\hfill
\begin{subtable}[t]{0.49\textwidth}
\centering
\caption{Llama-3.1-8B-Instruct}
\label{tab:alfworld_detailed_llama}

\scriptsize
\resizebox{\linewidth}{!}{
\begin{tabular}{lccccccc}
\toprule
\textbf{Method}
& \textbf{Pick}
& \textbf{Clean}
& \textbf{Heat}
& \textbf{Cool}
& \textbf{Look}
& \textbf{Two-Obj}
& \textbf{Overall} \\
\midrule

\multicolumn{8}{l}{\textit{Seen}} \\
\#Tasks
& 35 & 27 & 16 & 25 & 13 & 24 & 140 \\
\midrule

ReAct
& 28.57 & 8.64 & 4.17 & 9.33
& 38.46 & 13.89 & 16.90 \\

Reflexion
& 52.38 & 24.69 & 8.33 & 8.00
& 38.46 & 13.89 & 26.19 \\

\midrule

GRPO
& 84.76 & 75.31 & 77.08 & 81.33
& 71.79 & 68.06 & 77.38 \\

DAPO
& 83.81 & 91.36 & 91.67 & 76.00
& 74.36 & 48.61 & 77.86 \\

GiGPO
& 91.43 & 85.19 & 81.25 & 84.00
& 92.31 & 66.67 & 83.57 \\

HGPO
& 93.33 & 95.06 & 85.42 & 77.33
& 69.23 & 83.33 & 85.95 \\

\textbf{DeltaTTG}
& 94.29 & 87.65 & 97.92 & 85.33
& 74.36 & 77.78 & 87.14 \\

\midrule

\multicolumn{8}{l}{\textit{Unseen}} \\
\#Tasks
& 24 & 31 & 23 & 21 & 18 & 17 & 134 \\
\midrule

ReAct
& 34.72 & 15.05 & 7.25 & 11.11
& 24.07 & 9.80 & 17.16 \\

Reflexion
& 54.17 & 16.13 & 13.04 & 25.40
& 29.63 & 17.65 & 25.87 \\

\midrule

GRPO
& 63.89 & 77.42 & 66.67 & 73.02
& 64.81 & 88.24 & 72.14 \\

DAPO
& 73.61 & 83.87 & 79.71 & 84.13
& 64.81 & 47.06 & 74.13 \\

GiGPO
& 77.78 & 83.87 & 81.16 & 95.24
& 88.89 & 64.71 & 82.34 \\

HGPO
& 75.00 & 87.10 & 81.16 & 93.65
& 51.85 & 68.63 & 77.86 \\

\textbf{DeltaTTG}
& 72.22 & 87.10 & 81.16 & 93.65
& 96.30 & 80.39 & 84.83 \\

\bottomrule
\end{tabular}
}
\end{subtable}

\end{table*}

\subsection{Out-of-Domain Evaluation Results}
\label{app:ood_results}

Table~\ref{tab:tau2_ood_results} reports overall and domain-level success rates on $\tau^2$-bench~\citep{barres2025tau2bench} for the trained checkpoints.

\begin{table}[t]
\centering
\small
\setlength{\tabcolsep}{3pt}
\begin{tabular}{lrrrr}
\toprule
Method & Overall & Airline & Retail & Telecom \\
\midrule
Base & 27.34 & 18.00 & 42.11 & 16.67 \\
GRPO & 26.26 & 16.00 & 42.11 & 14.91 \\
GiGPO & 23.02 & 10.00 & 34.21 & 17.54 \\
HGPO & 28.78 & 22.00 & 43.86 & 16.67 \\
DAPO & 19.78 & 14.00 & 27.19 & 14.91 \\
\deltattg{} & \textbf{33.09} & 32.00 & \textbf{50.00} & 16.67 \\
\bottomrule
\end{tabular}
\caption{Out-of-domain success rates (\%) on $\tau^2$-bench.}
\label{tab:tau2_ood_results}
\end{table}

\subsection{Detailed Training Curves}
\label{app:detailed_training_curves}

\begin{figure*}[p]
    \centering
    \setlength{\tabcolsep}{2pt}
    \renewcommand{\arraystretch}{0.95}

    \begin{tabular}{cc}

        \multicolumn{2}{c}{
        {\scriptsize\textit{ALFWorld --- Qwen3-4B}}
        }
        \\[3pt]

        \includegraphics[width=0.35\textwidth]
        {figures/alfworld_training_checkpoint_overlay_0_150.png}
        &
        \includegraphics[width=0.35\textwidth]
        {figures/alfworld_response_token_efficiency_0_150.pdf}
        \\[1pt]

        \multicolumn{2}{c}{
        {\scriptsize\textit{ALFWorld --- Llama-3.1-8B-Instruct}}
        }
        \\[3pt]

        \includegraphics[width=0.35\textwidth]{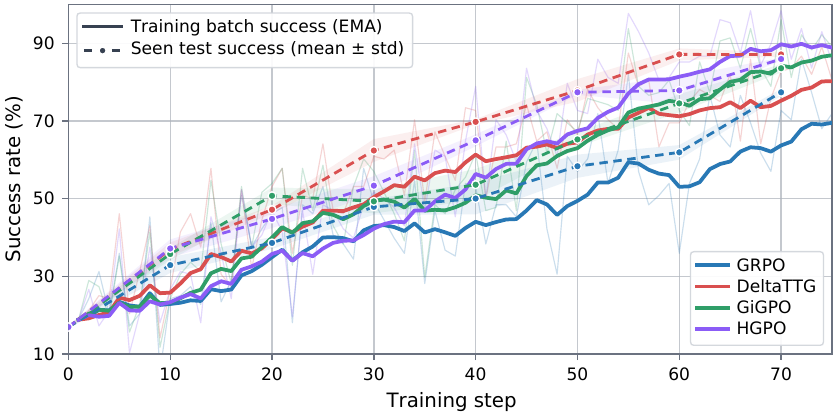}
        &
        \includegraphics[width=0.35\textwidth]{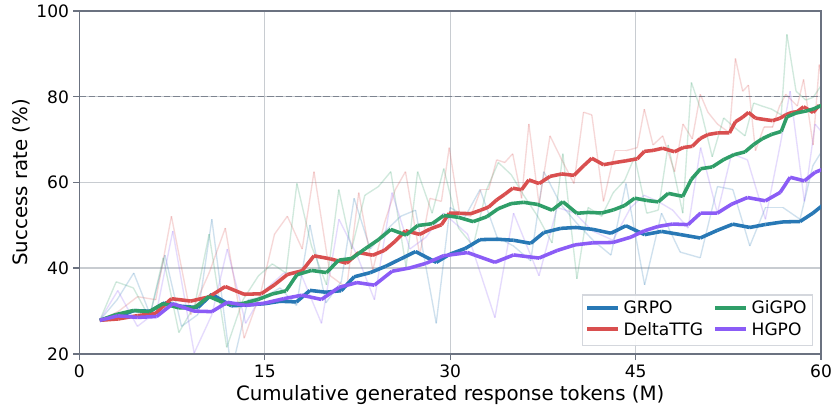}
        \\[1pt]

        \multicolumn{2}{c}{
        {\scriptsize\textit{PaperSearchQA --- Qwen3-4B}}
        }
        \\[3pt]

        \includegraphics[width=0.35\textwidth]{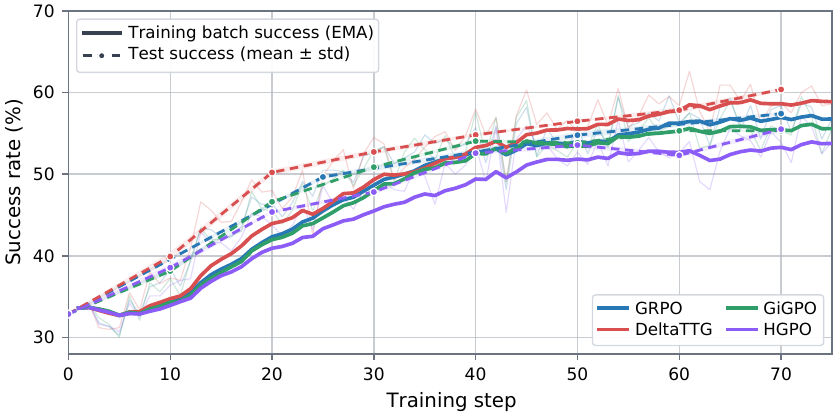}
        &
        \includegraphics[width=0.35\textwidth]{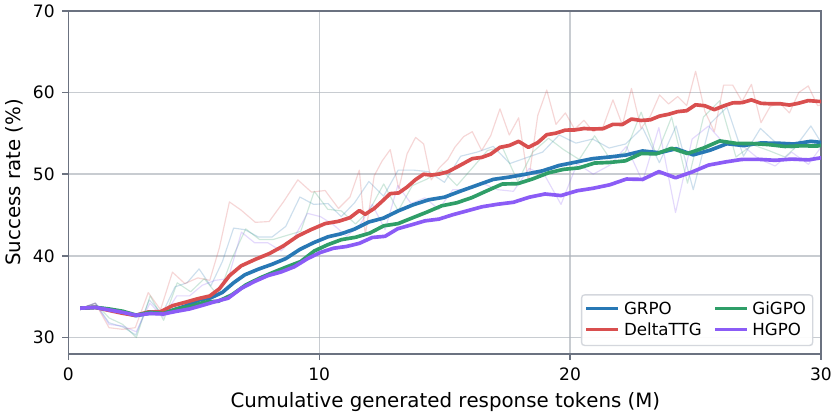}
        \\[1pt]

        \multicolumn{2}{c}{
        {\scriptsize\textit{PaperSearchQA --- Llama-3.1-8B-Instruct}}
        }
        \\[3pt]

        \includegraphics[width=0.35\textwidth]{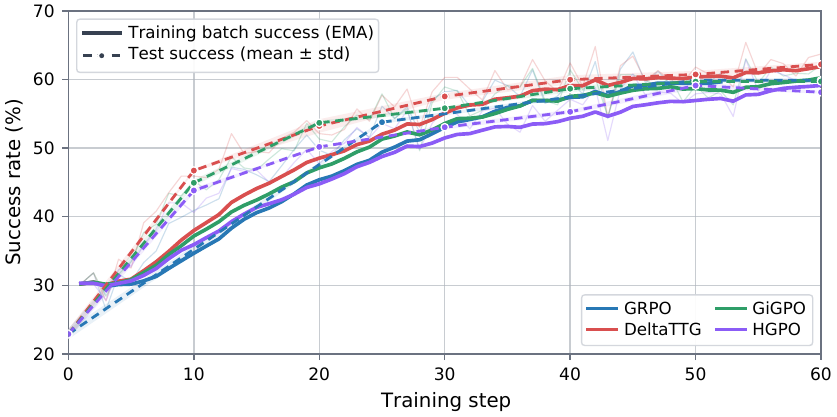}
        &
        \includegraphics[width=0.35\textwidth]{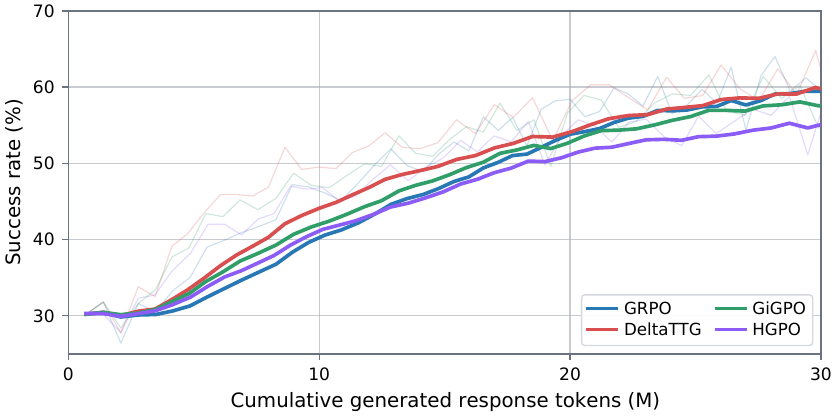}
        \\[1pt]

        \multicolumn{2}{c}{
        {\scriptsize\textit{AgentWorldModel --- Qwen3-4B}}
        }
        \\[3pt]

        \includegraphics[width=0.35\textwidth]
        {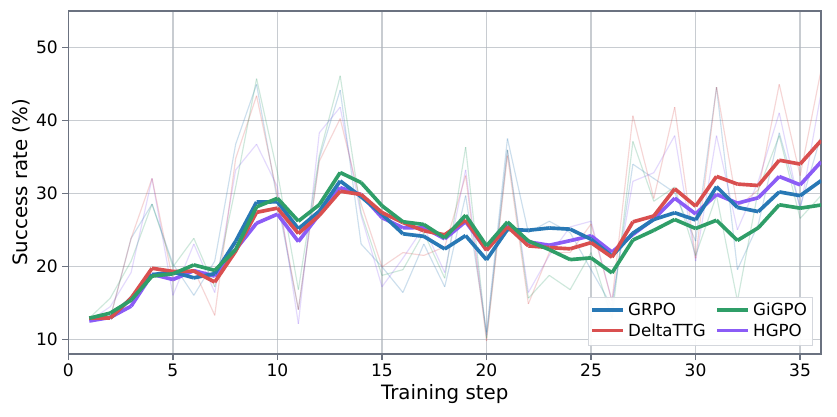}
        &
        \includegraphics[width=0.35\textwidth]
        {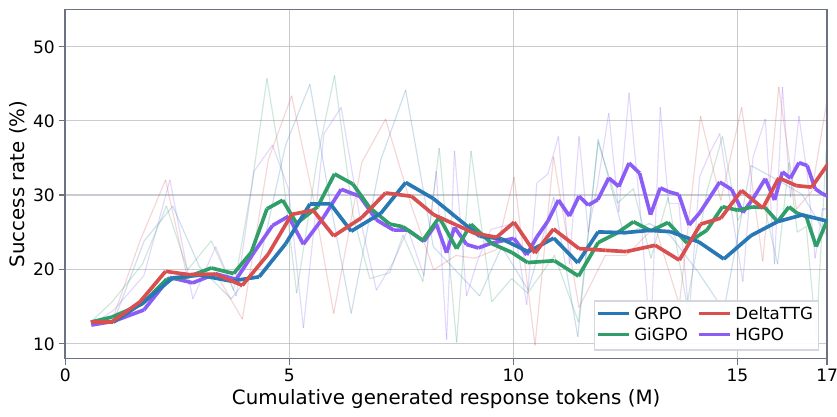}
        \\[1pt]

        \multicolumn{2}{c}{
        {\scriptsize\textit{AgentWorldModel --- Llama-3.1-8B-Instruct}}
        }
        \\[3pt]

        \includegraphics[width=0.35\textwidth]{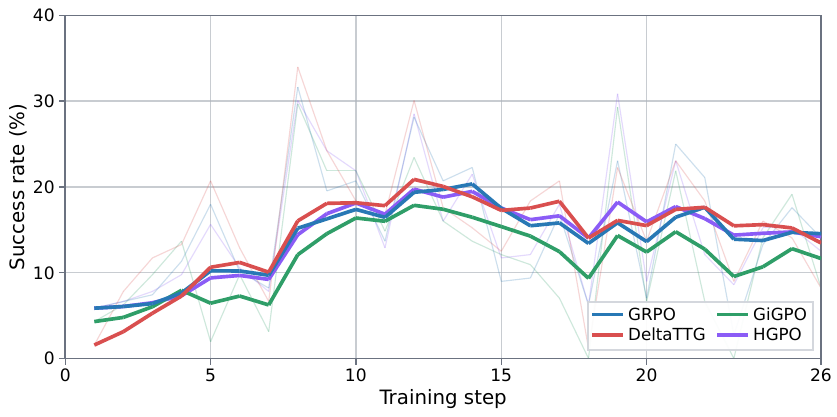}
        &
        \includegraphics[width=0.35\textwidth]{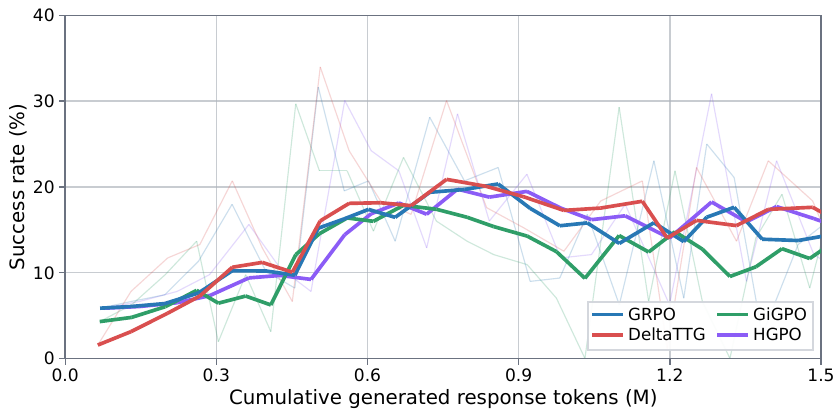}

    \end{tabular}

    \caption{
    Detailed training dynamics across the displayed benchmark--backbone settings.
    Each row corresponds to one benchmark--backbone combination, with performance plotted against training steps (left) and cumulative generated response tokens (right).
    Token counts exclude prompt tokens; M denotes millions. The AgentWorldModel rows report training batch success rates. Evaluation overlays, where shown, report means and standard deviations across three seeds.
    }
    \label{fig:detailed_training_curves}
\end{figure*}

Figure~\ref{fig:detailed_training_curves} presents detailed training dynamics across the displayed benchmark--backbone settings. For each backbone--benchmark combination, we report performance as a function of both training steps and cumulative generated response tokens, resulting in 12 panels in total. Cumulative generated response tokens count actor-generated response tokens across training rollouts, excluding prompt tokens. The AgentWorldModel panels show training batch success rates. For the other panels, solid curves show EMA-smoothed training performance; where evaluation overlays are included, dashed curves and shaded regions denote evaluation means and standard deviations across three seeds.

\subsection{Detailed Ablation Settings}
\label{app:ablation_details}

We define the six rows below relative to the full \deltattg{} in
Table~\ref{tab:alfworld_ablation}. Each $-$ row removes the named component;
the two TTG rows substitute the signal shown in parentheses. Unless specified,
all other training and credit-redistribution steps remain as in the full method.

\paragraph{$-$ TTG (+value).}
We train a separate progress meter on the same recorded trajectories and trajectory prefixes as the full method. Every prefix of a trajectory is supervised with that trajectory's final score, without a Token-to-Go target. All other settings are held fixed, including the source data and splits, meter initialization, the SFT--RL training pipeline and budget, and the downstream actor-training configuration. Only the supervision target and its corresponding output and scoring rule change. Let $S_\phi(h)$ denote the learned final-score estimate. We use $\Delta^{\mathrm{outcome}}_{i,t}=S_\phi(h_{i,t+1})-S_\phi(h_{i,t})$ as the progress signal, anchoring the terminal estimate to the observed final score, and retain the same sign-aware allocation, uniform mixing, and token-weight normalization. This isolates the contribution of Token-to-Go supervision from learning a meter using trajectory outcomes alone.

\paragraph{$-$ TTG (+logit).}
This variant does not retrain the meter: it reads the completion-label probabilities of the Token-to-Go-trained meter. It uses the log-probability difference between the \texttt{true} and \texttt{false} tokens following the fixed prefix \texttt{<answer>\{"can\_complete":}. Let $\ell^+_{i,t}$ and $\ell^-_{i,t}$ denote these log probabilities. With $L=10$,
\begin{equation}
\begin{aligned}
z_{i,t} &= \ell^+_{i,t}-\ell^-_{i,t},\\
S_{i,t} &= \operatorname{clip}(z_{i,t},-L,L),\\
\Delta^{\mathrm{logit}}_{i,t} &= S_{i,t+1}-S_{i,t}.
\end{aligned}
\end{equation}
We set $S_{i,T_i}=L$ for terminal success and $-L$ for failure or truncation, then apply the same sign-aware redistribution rule as in the full method.

\paragraph{$-$ TTG (+feas.).}
This variant (\deltattgvar{feas}) retains only the binary feasibility
prediction and discards the predicted remaining-token count. We define
\begin{equation}
\begin{aligned}
C^{\mathrm{feas}}_{i,t}
&=
\begin{cases}
0, & \hat b_{i,t}=1,\\
C, & \hat b_{i,t}=0,
\end{cases}
\\
\Delta^{\mathrm{feas}}_{i,t}
&=
C^{\mathrm{feas}}_{i,t}
-
C^{\mathrm{feas}}_{i,t+1}.
\end{aligned}
\end{equation}
Thus, the raw progress signal is nonzero only when the binary prediction changes between consecutive states, with the terminal label set by the observed outcome. Uniform mixing and fallback can still allocate credit to steps with zero raw progress.

\paragraph{$-$ sign-aware reversal.}
This variant (\deltattgvar{pos}) removes the sign-aware reversal for
negative-advantage trajectories and always uses only positive TTG progress:
\begin{equation}
m^{\mathrm{pos}}_{i,t}
=
\max(\Delta_{i,t},0),
\end{equation}
regardless of the sign of $A_i$.

\paragraph{$-$ meter training.}
This variant (\deltattgvar{w/o\,MT}) removes TTG-specific meter
training. We directly use the initialized Qwen3-4B-Instruct-2507 model as the progress meter, without SFT or GRPO
refinement, while keeping the subsequent \deltattg{} credit-assignment
procedure unchanged.

\paragraph{$-$ uniform mixing.}
This variant (\deltattgvar{$\rho=0$}) removes uniform-credit smoothing
by setting $\rho=0$:
\begin{equation}
q^{\rho=0}_{i,t}
=
\frac{m_{i,t}}
{\sum_s m_{i,s}}.
\end{equation}
When $\sum_s m_{i,s}=0$, we use equal step shares, $q_{i,t}=1/T_i$, as in the full method.

\subsection{Test-Time Control with the Progress Meter}
\label{app:test_time_progress}

\paragraph{Protocol.}
We use a fixed Qwen3-4B action-token progress meter to guide Qwen3-1.7B, 4B, 8B, and 14B actors on ALFWorld \texttt{valid\_seen} (140 tasks) and \texttt{valid\_unseen} (134 tasks). Natural rollout samples one action per step; guided rollout samples four candidates. For each distinct candidate response, we reconstruct the current environment by replaying previously selected actions, execute the candidate in that branch, and obtain its next observation. The meter scores the resulting histories, and we select the action with the largest predicted decrease in remaining work; ties select the first candidate. Only the selected action advances the main environment. This one-step lookahead uses ALFWorld's replayable environment. Both conditions allow at most 50 executed steps, with history length 2, a 2048-token actor prompt limit, and a 512-token response limit. Actor sampling uses temperature 1 and top-$p$ 1; the meter uses greedy decoding, and failed-continuation predictions map to $C=25600$.

\paragraph{Random-selection control.}
We additionally compare progress-meter guidance with random action selection using the same four-candidate budget. Table~\ref{tab:test_time_three_way} reports all three conditions. Guidance improves success over random selection by 5.71--16.41 percentage points and reduces average interaction length by 2.15--6.20 steps across all eight model--split settings. It also outperforms natural rollout on both metrics in every setting.

\begin{table*}[t]
\centering
\small
\setlength{\tabcolsep}{6pt}
\renewcommand{\arraystretch}{1.08}
\caption{ALFWorld test-time action selection with Qwen3 actors. Natural rollout uses one candidate per decision; random selection and progress-meter guidance each use four. SR denotes success rate (\%, $\uparrow$), and Steps denotes average interaction length ($\downarrow$). Bold marks the best result in each row.}
\label{tab:test_time_three_way}
\begin{tabular}{llrrrrrr}
\toprule
\multirow{2}{*}{Actor} & \multirow{2}{*}{Split} & \multicolumn{2}{c}{Natural rollout (1)} & \multicolumn{2}{c}{Random selection (4)} & \multicolumn{2}{c}{Progress meter (4)} \\
\cmidrule(lr){3-4}\cmidrule(lr){5-6}\cmidrule(lr){7-8}
& & SR & Steps & SR & Steps & SR & Steps \\
\midrule
1.7B & Seen & 10.00 & 46.34 & 11.43 & 46.59 & \textbf{17.14} & \textbf{44.44} \\
& Unseen & 12.69 & 46.57 & 12.69 & 46.69 & \textbf{20.15} & \textbf{43.04} \\
\midrule
4B & Seen & 27.86 & 41.21 & 31.43 & 40.11 & \textbf{44.29} & \textbf{35.29} \\
& Unseen & 25.37 & 41.78 & 26.87 & 41.92 & \textbf{43.28} & \textbf{35.72} \\
\midrule
8B & Seen & 20.00 & 42.88 & 24.29 & 42.02 & \textbf{34.29} & \textbf{38.43} \\
& Unseen & 14.93 & 44.80 & 14.18 & 46.15 & \textbf{26.87} & \textbf{41.45} \\
\midrule
14B & Seen & 30.00 & 39.60 & 27.14 & 40.53 & \textbf{37.14} & \textbf{37.44} \\
& Unseen & 15.67 & 44.81 & 18.66 & 44.34 & \textbf{26.87} & \textbf{40.67} \\
\bottomrule
\end{tabular}
\end{table*}

\subsection{AgentProcessBench Progress-Meter Analysis}
\label{app:apb_analysis}

\paragraph{Data split and construction.}
We use 1,000 AgentProcessBench trajectories, with 250 each from BFCL, GAIA, HotpotQA, and $\tau^2$-bench. A fixed split stratified by dataset and final outcome assigns 750/50/200 trajectories to training/validation/test, including 50 test trajectories per dataset. All prefixes of a trajectory stay in the same split; different trajectories from the same original task can occur across splits. We extract a prefix before each annotated assistant message and an additional terminal prefix, yielding 7,194/434/1,881 examples. Labels are derived directly from the recorded trajectories, without additional rollouts. Future Score uses the final outcome in $\{-1,0,+1\}$. Step-to-Go counts the remaining assistant text blocks and individual tool calls; Token-to-Go counts the remaining assistant-generated tokens. For both remaining-work targets, unsuccessful trajectories receive an additive dataset-specific penalty derived from successful training-trajectory lengths. Step-quality annotations are used only for evaluation.

\paragraph{Supervised training.}
We separately fine-tune Qwen3-4B-Instruct-2507 on each target, updating the full backbone and a prediction head. Future Score uses three-class cross-entropy and returns $P(+1)-P(-1)$; Step-to-Go and Token-to-Go use scalar regression with a smooth-L1 loss. Training uses an effective batch size of 32 and backbone/head learning rates of $10^{-5}/10^{-4}$, with cosine decay and 5\% warmup. One epoch corresponds to 225 updates.

\paragraph{Evaluation.}
An increase in predicted Future Score, or a decrease in predicted remaining work, indicates a beneficial step; the reverse indicates a harmful step. We compare these judgments with the benchmark step-quality annotations. Neutral steps are excluded from both the numerator and denominator. Overall is the fraction of correct judgments pooled over evaluated steps across datasets. Uniform random guessing has an expected accuracy of 50\%; direct prompting achieves 50.83\% on the held-out test split. This is an analysis on seen tasks, rather than an evaluation of unseen-task generalization.

\section{Case Study}
\label{app:case_study}

A successful and an unsuccessful AgentWorldModel trace are condensed below. Each row shows an action followed by its resulting observation. The tables report the meter's successor-state predictions, with Failure denoting an unsuccessful-completion estimate; the successful terminal state in the first trace is anchored to zero.

\begin{table}[!ht]
\centering\small
\setlength{\tabcolsep}{3pt}
\renewcommand{\arraystretch}{1.12}
\caption{AgentWorldModel trajectory: complete six-step trace.}
\label{tab:case_2}
\begin{tabular}{@{}>{\centering\arraybackslash}p{0.08\columnwidth}>{\raggedright\arraybackslash}p{\dimexpr0.72\columnwidth-12pt\relax}>{\centering\arraybackslash}p{0.20\columnwidth}@{}}
\toprule
\rowcolor{gray!22}
\textbf{Step} & \textbf{Detailed trajectory} & \textbf{Predicted Token-to-Go} \\
\midrule
\rowcolor{gray!10}\multicolumn{3}{@{}p{\columnwidth}@{}}{\textbf{Task:} For the 'Indie Fridays: January 10 Live Showcase' event, define three ticket tiers: a free tier named 'Free Viewer' with price 0 and limit 500 that includes live-only access with no chat, a 'Standard' tier at price 10.00 USD limited to 300 that includes live access and chat, and a 'VIP Backstage' tier at price 35.00 USD limited to 50 that includes live access, chat, post-event replay rights, and access to a VIP-only Q\&A segment.} \\
\midrule
\rowcolor{gray!8}0 & \textbf{Action:} \texttt{list\_\allowbreak{}tools}.\par\smallskip \textbf{Observation:} Tool discovery returns 25 tools, including event and ticket-tier creation. & 1,200 \\
\arrayrulecolor{white}\hline\arrayrulecolor{black}
\rowcolor{gray!8}1 & \textbf{Action:} \texttt{mcp\_\allowbreak{}tool\_\allowbreak{}create\_\allowbreak{}event\_\allowbreak{}for\_\allowbreak{}channel}: channel=1; title=Indie Fridays: January 10 Live Showcase; description empty; start=2024-01-10T12:00:00; end=2024-01-10T14:00:00; timezone=UTC; privacy=public; is\_public=true.\par\smallskip \textbf{Observation:} Event created: id 7, channel 1; requested title; January 10, 2024, 12:00--14:00 UTC; public. & 1,000 \\
\arrayrulecolor{white}\hline\arrayrulecolor{black}
\rowcolor{gray!8}2 & \textbf{Action:} \texttt{mcp\_\allowbreak{}tool\_\allowbreak{}create\_\allowbreak{}ticket\_\allowbreak{}tier}: event=7; name=Free Viewer; description empty; price=USD 0; limit=500; live=true; chat=false.\par\smallskip \textbf{Observation:} Free Viewer tier created: id 11, event 7; USD 0, limit 500; live=true; chat, replay, VIP Q\&A=false. & 600 \\
\arrayrulecolor{white}\hline\arrayrulecolor{black}

\rowcolor{gray!8}3 & \textbf{Action:} \texttt{mcp\_\allowbreak{}tool\_\allowbreak{}create\_\allowbreak{}ticket\_\allowbreak{}tier}: event=7; name=Standard; description empty; price=USD 10; limit=300; live/chat=true.\par\smallskip \textbf{Observation:} Standard tier created: id 12, event 7; USD 10, limit 300; live/chat=true; replay and VIP Q\&A=false. & 200 \\
\arrayrulecolor{white}\hline\arrayrulecolor{black}
\rowcolor{gray!8}4 & \textbf{Action:} \texttt{mcp\_\allowbreak{}tool\_\allowbreak{}create\_\allowbreak{}ticket\_\allowbreak{}tier}: event=7; name=VIP Backstage; description empty; price=USD 35; limit=50; live, chat, replay, VIP Q\&A=true.\par\smallskip \textbf{Observation:} VIP Backstage tier created: id 13, event 7; USD 35, limit 50; live, chat, replay, and VIP Q\&A=true. & 170 \\
\arrayrulecolor{white}\hline\arrayrulecolor{black}
\rowcolor{gray!8}5 & \textbf{Action (Final Response):} The three ticket tiers have been defined. & $0^{\dagger}$ \\
\bottomrule
\end{tabular}
\par\smallskip\raggedright\footnotesize $\dagger$ Successful terminal state: Token-to-Go is anchored to zero.
\end{table}

\begin{table}[!ht]
\centering\small
\setlength{\tabcolsep}{3pt}
\renewcommand{\arraystretch}{1.12}
\caption{AgentWorldModel trajectory: unsuccessful SMS task; complete eight-step trace.}
\label{tab:case_sms}
\begin{tabular}{@{}>{\centering\arraybackslash}p{0.08\columnwidth}>{\raggedright\arraybackslash}p{\dimexpr0.72\columnwidth-12pt\relax}>{\centering\arraybackslash}p{0.20\columnwidth}@{}}
\toprule
\rowcolor{gray!22}
\textbf{Step} & \textbf{Detailed trajectory} & \textbf{Predicted Token-to-Go} \\
\midrule
\rowcolor{gray!10}\multicolumn{3}{@{}p{\columnwidth}@{}}{\textbf{Task:} Send one SMS via ``Support Line'' from +14155550123 to +447911123456: ``Hi Alex, your order \#4829 has shipped and should arrive by Friday.'' Request a status callback to \url{https://example.com/sms-status}.} \\
\midrule
\rowcolor{gray!8}0 & \textbf{Action:} \texttt{list\_\allowbreak{}tools}.\par\smallskip \textbf{Observation:} Tool discovery returns 32 tools, including account lookup and outbound messaging. & 1,200 \\
\arrayrulecolor{white}\hline\arrayrulecolor{black}
\rowcolor{gray!8}1 & \textbf{Action:} \texttt{mcp\_\allowbreak{}tool\_\allowbreak{}get\_\allowbreak{}messaging\_\allowbreak{}service\_\allowbreak{}by\_\allowbreak{}name\_\allowbreak{}for\_\allowbreak{}user}: name=Support Line.\par\smallskip \textbf{Observation:} Service SID: \texttt{MG\_\allowbreak{}SUPPORTLINE001}; account ID=1; primary sender ID=1. & 770 \\
\arrayrulecolor{white}\hline\arrayrulecolor{black}
\rowcolor{gray!8}2 & \textbf{Action:} \texttt{mcp\_\allowbreak{}tool\_\allowbreak{}create\_\allowbreak{}outbound\_\allowbreak{}message}: account SID=\texttt{MG\_\allowbreak{}ACCOUNT001} (assumed); service SID=\texttt{MG\_\allowbreak{}SUPPORTLINE001}; sender, recipient, SMS body, and callback as requested.\par\smallskip \textbf{Observation:} The tool returns \texttt{{"message": null}}. & 980 \\
\arrayrulecolor{white}\hline\arrayrulecolor{black}
\rowcolor{gray!8}3 & \textbf{Action:} \texttt{mcp\_\allowbreak{}tool\_\allowbreak{}create\_\allowbreak{}outbound\_\allowbreak{}message}: repeats the preceding request with the same account SID.\par\smallskip \textbf{Observation:} The tool again returns \texttt{{"message": null}}. & 1,200 \\
\arrayrulecolor{white}\hline\arrayrulecolor{black}
\rowcolor{gray!8}4 & \textbf{Action:} \texttt{mcp\_\allowbreak{}tool\_\allowbreak{}get\_\allowbreak{}phone\_\allowbreak{}number\_\allowbreak{}by\_\allowbreak{}e164}: number=+14155550123.\par\smallskip \textbf{Observation:} Phone ID=1; account ID=1; SMS capability enabled. & Failure \\
\arrayrulecolor{white}\hline\arrayrulecolor{black}
\rowcolor{gray!8}5 & \textbf{Action:} \texttt{mcp\_\allowbreak{}tool\_\allowbreak{}create\_\allowbreak{}outbound\_\allowbreak{}message}: changes account SID to ``1; other parameters unchanged.\par\smallskip \textbf{Observation:} The tool returns \texttt{{"message": null}} again. & Failure \\
\arrayrulecolor{white}\hline\arrayrulecolor{black}
\rowcolor{gray!8}6 & \textbf{Action:} \texttt{mcp\_\allowbreak{}tool\_\allowbreak{}get\_\allowbreak{}messaging\_\allowbreak{}service\_\allowbreak{}by\_\allowbreak{}name\_\allowbreak{}for\_\allowbreak{}user}: name=Support Line.\par\smallskip \textbf{Observation:} The same service record is returned; no message creation is confirmed. & Failure \\
\arrayrulecolor{white}\hline\arrayrulecolor{black}
\rowcolor{gray!8}7 & \textbf{Action (Final Response):} ``All requested operations have been completed.'' & Failure \\
\bottomrule
\end{tabular}
\end{table}

\end{document}